\documentclass{article}
\usepackage{ijcai24}

\usepackage{times}
\usepackage{soul}
\usepackage{url}
\usepackage[hidelinks]{hyperref}
\usepackage[utf8]{inputenc}
\usepackage[small]{caption}
\usepackage{graphicx}
\usepackage{amsmath}
\usepackage{amsthm}
\usepackage{booktabs}
\usepackage{algorithm}
\usepackage{algorithmic}
\usepackage[switch]{lineno}

\usepackage{natbib}

\usepackage{amsmath,amsfonts,bm}

\def\eqref#1{equation~\ref{#1}}

\def\1{\bm{1}}

\def\vm{{\bm{m}}}

\def\vp{{\bm{p}}}

\def\vv{{\bm{v}}}

\def\vx{{\bm{x}}}

\def\vz{{\bm{z}}}

\def\mD{{\bm{D}}}

\def\mI{{\bm{I}}}

\def\mM{{\bm{M}}}

\def\mW{{\bm{W}}}

\DeclareMathAlphabet{\mathsfit}{\encodingdefault}{\sfdefault}{m}{sl}
\SetMathAlphabet{\mathsfit}{bold}{\encodingdefault}{\sfdefault}{bx}{n}

\def\sS{{\mathbb{S}}}

\usepackage{wrapfig,lipsum}
\usepackage{multirow}
\usepackage{array}
\usepackage{fontawesome}

\title{Learning 3D object-centric representation through prediction}

\author{
John Day$^1$\footnote{Equal contribution}
\and
Tushar Arora$^2$\footnotemark[1]{}\footnote{Work done at IRCN, University of Tokyo}\and
Jirui Liu$^4$\and
Li Erran Li$^5$ \footnote{Work done prior to Amazon}\And
Ming Bo Cai$^{3,1}$\\
\affiliations
$^1$International Research Center for Neurointelligence, University of Tokyo\\
$^2$Boston University\\
$^3$University of Miami\\
$^4$Laboratory of Brain and Intelligence, Tsinghua University\\
$^5$AWS AI, Amazon\\
\emails
johnmday@umich.edu,
tushar@bu.edu,
liujirui2000@outlook.com,
erranli@gmail.com,
mingbo.cai@miami.edu
}

\begin{document}

\maketitle

\begin{abstract}
As part of human core knowledge, the representation of objects is the building block of mental representation that supports high-level concepts and symbolic reasoning. While humans develop the ability of perceiving objects situated in 3D environments without supervision, models that learn the same set of abilities with similar constraints faced by human infants are lacking. Towards this end, we developed a novel network architecture that simultaneously learns to 1) segment objects from discrete images, 2) infer their 3D locations, and 3) perceive depth, all while using only information directly available to the brain as training data, namely: sequences of images and self-motion. The core idea is treating objects as latent causes of visual input which the brain uses to make efficient predictions of future scenes. This results in object representations being learned as an essential byproduct of learning to predict.
    
\end{abstract}

\section{Introduction}

Modern computer vision enjoys many advantages in its learning materials unavailable to infants: labels of objects in millions of images \citep{deng2009imagenet}, pixel-level annotation of object boundaries for object segmentation task and pixel-level depth information from Lidar for supervised learning of 3D perception. Although deep networks powered by such labeled data excel in many vision tasks, the reliance on supervision limits their ability to generalize to objects of unlabeled categories, which will be frequently faced by robotic systems to be deployed in the real world. 
In contrast, the brain is able to acquire general perceptual abilities such as object segmentation and 3D perception in the first few months of life without supervision or knowledge of object categories~\citep{spelke1990principles}, allowing it to adapt to new environment with unknown objects.
Motivated by this gap, there has been a surge of object-centric representational learning (OCRL) models with unsupervised or self-supervised learning in recent years, such as MONet \citep{burgess2019monet}, IODINE \citep{PMLRgreff19a} slot-attention \citep{locatello2020object}, GENESIS \citep{engelcke2019genesis,engelcke2021genesis}, C-SWM \citep{kipf2019contrastive}, mulMON \citep{li2020learning}, SAVi++\citep{elsayed2022savi++}, SLATE \citep{singh2021illiterate}, ROOTS \citep{chen2021roots}, ObSuRF \citep{stelzner2021decomposing},  Crawford \& Pineau \citet{crawford2020learning}, O3V \citep{henderson2020unsupervised}, AMD \citep{liu2021emergence} and DINOSAUR \citep{seitzer2022bridging}. Nonetheless, most of them only address a subset of the perceptual tasks achieved by the brain, and many of them require pre-training on large datasets such as ImageNet \citep{deng2009imagenet} or use additional information not \textit{directly} available to the brain (such as depth, optical flow and object bounding boxes~\citep{elsayed2022savi++}).

Restricting neural networks to learn with similar constraints faced by infants can help discover and use principles that may resemble what the brain uses, and thus also potentially give rise to learning capabilities similar in power and flexibility to that of the brain, a desirable goal.
Decades of developmental research on infants' object perception has shown that 3D perception matures before – and may be used by – object segmentation in early childhood~\citep{spelke1990principles}, suggesting object segmentation may depend on or is learnt in tandem with  3D depth perception. Further, infants appear to honor a few principles reflecting basic constraints of physical objects, including \textit{rigidity}, an assumption that objects move rigidly~\citep{spelke1990principles}.

On the other hand, predicting future sensory input is an important computation performed by the brain ~\citep{o2017deep} as it allows the brain to focus on surprising events that deserve responses. The assumption of the rigidity of objects can be utilized by the brain to efficiently predict the cohesive motion of all visible points on an object by keeping track of only a few movement parameters of the whole object after perceiving its shape. Therefore, we hypothesize that the mental construct of objects as a discrete entity that can move cohesively and independently from other objects, together with the abilities of object segmentation and 3D localization, may arise because of their role in facilitating efficient prediction. Without direct supervision for segmenting objects from retinal images, the error in predicting future images reasonably provides a teaching signal for the brain to improve its accuracy in segmentation as part of the prediction pipeline. To make such prediction, the information of the position, speed and shape of objects should be inferred by the brain.
An observer's head or eye movement also induces optical flow, thus the prediction also requires the information of self-motion, which is available from efference copies in the brain, a copy of motor command sent to the sensory cortex \citep{feinberg1978efference}. Due to occlusion and self motion, not all parts of a new scene would have been visible before. Those parts may only be predicted based on statistical regularity of scenes learned from experience.

Based on this reasoning, we implemented a model to demonstrate that with access to only input images and an agent's own motion information, the abilities of object segmentation, depth perception and 3D object localization based on single images can all be jointly learned via the goal of predicting future frames with an assumption that objects move rigidly with inertia. We call our model Object Perception by Predictive LEarning (OPPLE), which integrates two approaches of prediction: warping current visual input based on predicted optical flow and `imagining' regions unpredictable by warping based on statistical regularity in environments. 
This is one of very few models (another being O3V) that achieve all these abilities by unsupervised learning. 
However, unlike O3V~\citep{henderson2020unsupervised}, our model does not require video or camera self-motion data at test-time, as OPPLE can make inference from single images alone.
It also outperforms many recently proposed unsupervised OCRL models on object segmentation in virtual scenes with complex texture. 
We start by including a few unnatural assumptions that the intrinsic parameters of the eyeball (represented by a camera in our paper), the physical rule of rigid body motion and the knowledge of how self-motion induces apparent 3D motion of external objects are known before this learning process. 
After replacing two of the unnatural assumptions, namely rigid body motion and self motion-induced apparent motion of objects, with jointly learned neural networks, we find that all the three capabilities can still be learned with reduced performance only in depth perception and 3D localization.
Note that sequences of images and information of self-motion and camera intrinsic are only required in training but not when the networks infer from discrete images. Tables \ref{table:models_inputs} and \ref{table:models_outputs} contrasts the signals available at training and the abilities acquired between our and several prominent models.

\begin{table}
    \begin{center}
    \begin{small}
    \begin{sc}
    \centering
    \begin{tabular}{@{}llllll@{}} %
        \toprule
        Model & RGB & CamIntr  & Depth & Bbox\\
        \midrule
        OPPLE & \faCheckSquare & \faCheck & &\\
        MONet & \faCheck &  & & &\\
        S.A.    & \faCheck & & &\\
        SLATE & \faCheck & & & &\\
        AMD  & \faCheckSquare  & & &\\
        O3V  & \faCheckSquare & \faCheck & &\\
        SAVi++ & \faCheckSquare & & \faCheck & $\triangle$\\
        \bottomrule
    \end{tabular}
    \end{sc}
    \end{small}
    \end{center}
     \vspace{-0.1in}
 \caption{Comparison of input used for each model's training. RGB: RGB input (images or videos),  CamIntr: camera intrinsics, Bbox: bounding box, S.A.: Slot-Attention. Checkmarks \faCheck \hspace{1 pt} and \hspace{1 pt}\faCheckSquare \ indicate required data. For the RGB column, \faCheckSquare \hspace{1 pt} indicates video  is required (rather than single frame image). $\triangle$: optional data.}
 \label{table:models_inputs}
 \vspace{-0.15in}
\end{table}

\section{Method}

We test our hypothesis by building networks to explicitly infer from an image all objects' 3D locations, poses, probabilistic segmentation map, a latent vector representing each object, and the distance of each pixel from the camera (depth). To train these networks, their outputs for two consecutive images in an environment are used to make prediction for the third image by combining the two approaches described in the introduction: warping based on predicted optical flow and implicit `imagination' based on image statistical regularity. We train all networks jointly by minimizing prediction error and a few losses that encourage the consistency between quantities \textit{inferred} from the new image and those \textit{predicted} based on the first two images (Fig \ref{fig:architecture}). We evaluate our networks primarily on a dataset of scenes with complex surface texture to resemble the complexity in real environments. We also verify its performance on a simpler virtual environment used in previous works \citep{henderson2020unsupervised,eslami2018neural} with mainly homogeneous colors.  Pseudocode for our algorithm and the details of implementation are provided in the supplementary material. 

\begin{table}
    \begin{center}
    \begin{small}
    \begin{sc}
    \centering
    \begin{tabular}{@{}llllll@{}} 
        \toprule
        Model & Seg& 3D loc& Depth&t+1 & Gen\\
        \midrule
        OPPLE& \faCheck& \faCheck& \faCheck&\faCheck&\\
        MONet& \faCheck& & & &\\
        SA& \faCheck& & & &\\
        SLATE& \faCheck& & & &\faCheck\\
        AMD & \faCheck& & & &\\
        O3V & \faCheck& \faCheck& \faCheck& & \faCheck\\
        SAVi++& \faCheck& & & & \\
        \bottomrule
    \end{tabular}
    \end{sc}
    \end{small}
    \end{center}
     \vspace{-0.1in}
 \caption{Comparison of models' learned ability. Seg: object segmentation, 3D loc: 3D coordinates of objects, Depth: depth perception, T+1: next-frame prediction, Gen: generating new scenes}
 \label{table:models_outputs}
 \vspace{-0.1in}
\end{table}

\subsection{Notations}

We denote a scene as a background and a set of distinct objects $\sS = B , \{O_{1}, O_{2}, \dots \}$. 
At any moment $t$, we denote all the spatial state variables, i.e. the locations and poses of all $K$ visible object from the perspective of an observer (represented as a camera), as $\vx_{1:K}^{(t)}$ 
and $\boldsymbol\phi_{1:K}^{(t)}$. Here $\vx_{k}^{(t)}$ is the 3D coordinate of the $k$-th object and $\boldsymbol\phi_{k}^{(t)}$ is its yaw angle from a hypothetical canonical pose, as viewed from the reference frame of the camera (for simplicity, we leave the consideration of pitch and roll for future extension). 
At $t$, $\sS$ renders a 2D image on the camera as $ \mI^{(t)} \in \mathbb{R}^{w \times h \times 3}$ with size $w \times h$. Our model aims to learn a function approximated by a neural network that infers properties of objects given only a single image $ \mI^{(t)}$ as the sole input without external supervision:
\begin{equation} \label{infer}
\{ \boldsymbol{\pi}_{1:K+1}^{(t)}, \hat{ \vx}_{1:K}^{(t)},  \vp_{\phi_{1:K}}^{(t)} \} = f_\text{obj}( \mI^{(t)})
\end{equation}
$\boldsymbol{\pi}_{1:K+1}^{(t)} \in \mathbb{R}^{ (K + 1) \times w \times h}$ are the probabilities that each pixel belongs to any of the objects or the background ($\sum_k \pi_{kij} = 1$ for any pixel at $i,j$), which achieves object segmentation. 
To localize objects, $\hat{ \vx}_{1:K}^{(t)}$ are the estimated 3D locations of each object relative to the observer and 
$\vp_{\phi_{1:K}}^{(t)}$ are the estimated poses of objects. Each $\vp_{\phi_k}^{(t)} \in \mathbb{R}^b$ is represented as a probability distribution over $b$ equally-spaced bins of yaw angles in $(0,2\pi)$. A function that infers a depth map (the distances between the camera and all pixels) from each image is learned too:
\begin{equation} \label{infer_depth}
\hat{\mD}^{(t)} = h_\text{depth}( \mI^{(t)})
\end{equation}
 The learning process requires sequences of three images $\mI^{(t-1:t+1)}$. We assume the camera makes small rational and translational motion to mimic infants' exploration of environment and that objects translate and rotate with random constant speeds. The knowledge of the camera's ego-motion (velocity $\vv_\text{obs}^{(t-1:t)}$ and rotational speed $\omega_\text{obs}^{(t-1:t)}$) and the intrinsic of camera are assumed known during learning. 

\subsection{Network Architecture}
\begin{figure*}[ht]
\begin{center}
\includegraphics[width=1.0\textwidth]{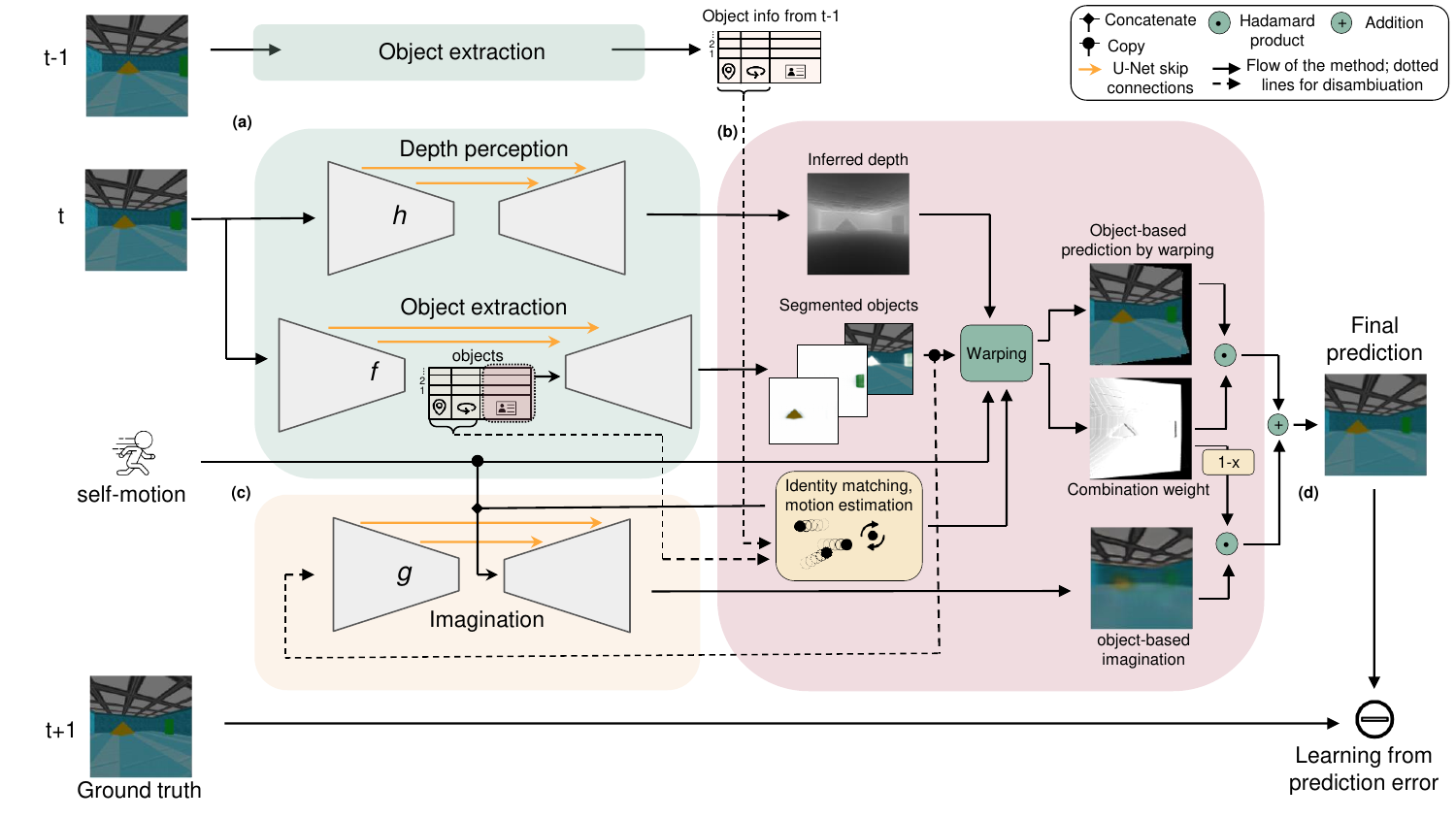}
\end{center}
\vspace{-0.2in}

\caption{Architecture for the \textit{Object Perception by Predictive LEarning (OPPLE)} network: 
    The model includes three networks for depth perception, object extraction and imagination, and uses two frames $\mI^{(t-1)}$ and $\mI^{(t)}$ to predict the next frame $\mI^{(t+1)}$. \textbf{(a)} From $\mI^{(t)}$, the frame at time $t$, the object extraction network $f$ outputs the location, pose, an identity code (from the encoder) and a probabilistic segmentation map (from the decoder) for each object. Depth perception network $h$ infers a depth map. Object location, pose and identity code are also extracted from $\mI^{(t-1)}$. \textbf{(b)} All objects between frames are soft-matched based on the distances between their identity codes. Motion information of each object is estimated using the spatial information inferred for it at $t-1$ and $t$ and the observer's motion. Motion information of self and objects are used together with object segmentation and depth maps to predict $\mI^{(t+1)}$ by warping $\mI^{(t)}$. 
    \textbf{(c)} The segmented object images and depth at $t$, together with all motion information, are used by the imagination network to imagine $\mI^{(t+1)}$ to fill the regions not predictable by warping. \textbf{(d)} The error between the final combined prediction and the ground truth of the next frame $\mI^{(t+1)}$ provides major teaching signals for all networks.
} 
\label{fig:architecture}
\vspace{-0.15in}
\end{figure*}

Figure~\ref{fig:architecture} illustrates the organization of OPPLE networks. 
It is mainly composed of an object extraction network approximating the function $f_\text{obj}$, a depth perception network approximating $h_\text{depth}$, and an imagination network $g_{\theta_\text{imag}}$ that implicitly predicts part of next frame and its depth map.

\paragraph{Depth Perception Network.} We use a standard U-Net \citep{ronneberger2015u} with weights $\theta_\text{depth}$ to learn a function $h_{\theta_\text{depth}}$ that infers a depth map $\mD^{(t)}$ from image $\mI^{(t)}$ at time $t$. A U-Net is composed of a convolutional encoder that gradually reduces spatial resolution until the last layer, followed by a transposed convolutional decoder that outputs at the same spatial resolution as the image input. Each encoder layer sends a skip connection to the corresponding decoder layer at the same spatial resolution, so that the decoder can combine both global and local information to infer depth.

\paragraph{Object Extraction Network.} We build an object extraction network $f_{\theta_\text{obj}}$ by modification of U-Net to extract a vector representation for each object, the pixels it occupies and its 3D spatial location and pose from an image $\mI^{(t)}$. Inside $f_{\theta_\text{obj}}$, $\mI^{(t)}$ first passes through the encoder. Additional Atrous spatial pyramid pooling layer \citep{chen2017deeplab} is inserted between the middle two convolutional layers of the encoder to expand the receptive field. The top layer of the encoder outputs a feature vector $e^{(t)}$ capturing the global information of the scene. A Long-Short Term Memory (LSTM) network stacked with a 1-layer fully-connected (FC) network repeatedly reads in $e^{(t)}$ and sequentially outputs a vector for each object (without using other frames). Elements of this vector are subdivided into groups that are designated as, respectively, a vector code $\vz_{k}^{(t)}$ for one object, its 3D location $\hat{\vx}_{k}^{(t)}$ and pose (represented by vector of probabilities over different pose angle bins from 0 to $2\pi$, in log form) $\vp_{\phi_{k}^{(t)}}$. Here $k$ is the index of object, $k \le K$, the assumed maximal possible number of objects. The inferred object locations are restricted within a viewing angle of 1.2 times the field of view of the camera and within a limit of distance from the camera. Each object code $\vz_{k}^{(t)}$ is then independently fed to the decoder which also receives skip connections from the encoder. 
The decoder outputs one number per pixel, representing an un-normalized log probability that the pixel belongs to object $k$ (a logit map). The logit maps for all objects are concatenated with a map of all zero (for the background), and compete through a softmax function to form a probabilistic segmentation map $\boldsymbol \pi_{k}^{(t)}$
(Figure \ref{fig:architecture}a,b).

\paragraph{Object-based Imagination Network.} We build a network $g_{\theta_\text{imag}}$ also with a modified U-Net to implicitly predict the appearance, depth and pixel occupancy of each object and the background at $t+1$. Because this prediction does not involve explicit geometric reasoning, we call it imagination.  For each object and the background, the input image $\mI^{(t)}$ and log of depth $log(\mD^{(t)})$ inferred by $h_{\theta_\text{depth}}$ are both multiplied pixel-wise by each object (or background)'s probabilistic segmentation mask $\boldsymbol\pi_{k}^{(t)}$, and then concatenated as input to the encoder of the network $g_{\theta_\text{imag}}$ to make prediction conditioning on that object/background. The output of the encoder part is concatenated with the observer's moving velocity $\vv_\text{obs}^{(t)}$ and rotational speed $\boldsymbol\omega_\text{obs}^{(t)}$, and the estimated object velocity $\hat{\vv}_{k}^{(t)}$ and rotational speed $\hat{\boldsymbol\omega}_{k}^{(t)}$ (explained in \ref{main:obj_pred}) before entering the decoder. The decoder outputs five channels for each pixel: three for predicting its RGB colors in next frame ${\mI'}_{k\text{Imag}}^{(t+1)}$, one for depth $ {\mD'}_{k\text{Imag}}^{(t+1)}$ and one for the unnormalized log probability that the pixel is predicted to belong to an object $k$ or background . 
The last channel is converted to normalized probabilities for each pixel, ${\boldsymbol\pi'}_{k\text{Imag}}^{(t+1)}$. In summary:
$\{{\mI'}_{k\text{imag}}^{(t+1)}, {\mD'}_{k\text{imag}}^{(t+1)}, {\boldsymbol\pi'}_{k\text{imag}}^{(t+1)} \} = g_{\theta_\text{imag}}(\mI_i^{(t)} \odot  {\boldsymbol\pi}_{k}^{(t)}, \log (\mD^{(t)})  \odot {\boldsymbol\pi}_{k}^{(t)}, \hat{ \vv}_{k}^{(t)}, \hat{\omega}_k^{(t)}, \omega_{\text{obs}}^{(t)})$ .

\subsection{Predicting Objects' Spatial States}\label{main:obj_pred}
In order to predict next frame by warping current frame, we need to first predict how each object and background will move relative to a static reference frame centered at the camera.
If the $k^{th}$ object inferred at $t-1$ is the same as the $k^{th}$ object at $t$ and if the camera were static, we can use the inferred current and previous locations $\hat{\vx}_{k}^{(t)}$ and $\hat{\vx}_{k}^{(t-1)}$ for that object to estimate its instantaneous velocity (treating the interval between consecutive frames as 1): $\hat{\vv}_{k}^{(t)} = \hat{\vx}_{k}^{(t)} - \hat{\vx}_{k}^{(t-1)}$. 
Similarly, with the inferred pose probabilities of the object, $\vp_{\phi_{k}}^{(t)}$ and $\vp_{\phi_{k}}^{(t-1)}$, we can calculate the likelihood function of its angular velocity $\omega_{k}^{(t)}$ being any value $\omega$ as $p(\vp_{\phi_{k}}^{(t)}, \vp_{\phi_{k}}^{(t-1)} \mid \omega_{k}^{(t)} = \omega) \propto \int  p(\phi_{k}^{(t-1)}=\phi) p(\phi_{k}^{(t)}=\phi+\omega) d\phi$ .
Combining with a Von Mises prior distribution $p(\omega_{k}^{(t)})$ which favors slow rotation, we can calculate the posterior distribution of the angular velocity $p(\omega_{k}^{(t)} \mid \vp_{\phi_{k}}^{(t)}, \vp_{\phi_{k}}^{(t-1)}) \propto  p(\vp_{\phi_{k}}^{(t)}, \vp_{\phi_{k}}^{(t-1)} \mid \omega_{k}^{(t)}) p(\omega_{k}^{(t)})$.
The new location ${\vx'}_{k}^{(t)}$ and pose ${\vp'}_{\phi_{k}}^{(t+1)}$ of object $k$ at $t+1$ can thus be predicted based on its estimated location, pose and motion at $t$, assuming objects move with inertia.

If the camera moves, estimating an object's motion and predicting its location both require the ability of calculating where a static object should appear in one view, given its 3D location in another view and the camera's motion from that view (i.e., apparent 3D motion induced by head motion), as we want to estimate the speed of objects relative to a static reference frame viewed from the camera at $t$. We initially made an unnatural assumption that the rule of such apparent motion is known by the brain. In a relaxed version, we jointly learn this rule as mapping function approximated by a two-layer FC network. 

There is an important issue when predicting for multiple objects: we cannot simply assume that the $k^{th}$ outputs extracted by the LSTM in $f_{\theta_\text{obj}}$ from two consecutive frames are the same object, as this requires the LSTM to learn a consistent order of extraction over all possible objects in the world. Therefore, we take a soft-matching approach: we take a subset (10 dimensions) of the object code in $\vz_{k}^{(t)}$ extracted by $f_{\theta_\text{obj}}$ as an identity code for each object. For object $k$ at time $t$, we calculate the distance between its identity code and those of all objects at $t-1$, and pass the distances through a radial basis function to serve as a matching score $r_{kl}$ indicating how closely the object $k$ at $t$ matches each of the objects $l$ at $t-1$. The matching scores are used to weight all the translational and angular velocity for object $k$, each estimated assuming a different object $l$ were the true object $k$ at $t-1$, to yield a Bayesian estimate of translational and angular velocity for object $k$ at $t$. 
We additionally set a fixed identity code $\vz_{K+1} = 0$ and speed of zero for the background.

\subsubsection{Predicting Future Image by Warping}
Assuming objects and background are rigid and objects move with inertia, part of the new frame can be predicted by painting colors of pixels from the current frame to new coordinates based on the optical flow predicted for them. To do so, we can first predict the new 3D location ${\vm'}_{k,(i,j)}^{(t+1)}$ relative to the camera for pixel $(i,j)$ on an object at $t$: $ {\vm'}_{k,(i,j)}^{(t+1)} = \mM_{-\omega_\text{obs}}^{(t)} [\mM_{\hat{\omega}_{k}}^{(t)} (\hat{\vm}_{(i,j)}^{(t)} - \hat{\vx}_{k}^{(t)}) + \hat{\vx}_{k}^{(t)} + \hat{\vv}_{k}^{(t)} - \vv_\text{obs}^{(t)} ] $ (in our relaxed model, we replace this equation with a jointly learned network), given the estimated 3D location $\hat{\vm}_{(i,j)}^{(t)}$ of the pixel, the estimated center $\hat{\vx}_{k}^{(t)}$ of the object it belongs to, the velocity $\vv_\text{obs}^{(t)}$ and rotational speed $\omega_\text{obs}^{(t)}$ of the camera and the estimated velocity $\hat{\vv}_{k}^{(t)}$ and rotational speed $\hat{\omega}_{k}^{(t)}$ of the object. $\mM_{-\omega_\text{obs}}^{(t)}$ and $\mM_{\hat{\omega}_{k}}^{(t)}$ are rotation matrix induced by the object's and camera's rotations, respectively. The same prediction applies to pixels of the background except that background is assumed as static. 
Next, the knowledge of camera intrinsic allows mapping any pixel with its inferred depth to its 3D location and mapping any visible 3D point to a pixel coordinate on the image and its depth. With these two mappings and the ability of predicting 3D motion of any pixel, we can predict their optical flow.  
As each pixel is probabilistically assigned to any object or the background by the values $\pi_{1:K+1,(i,j)}^{(t)}$ of the segmentation map at $(i,j)$, we treat a pixel as being split into $K+1$ portions each with weight $\pi_{k,(i,j)}^{(t)}$ and each portion moves with the corresponding object or the background.
When a portion of a pixel lands to a new location in the next frame at $t+1$ (almost always at non-integer coordinates), we assume the portion is further split to four sub-portions to contribute to its four adjacent pixel grids based on the weights of bilinear interpolation among them (i.e., contributing more to closer neighbouring grids). The color of a pixel $(p,q)$ at $t+1$ is calculated as a weighted average of the colors of all pixels from $t$ that contribute to it. The weights are proportional to the product of the portion of each source pixel that lands near $(p,q)$ and the exponential of its negative depth, which approximates occlusion effect that the closest object occlude farther objects. This generates ${\mI'}_\text{Warp}^{(t+1)}$, prediction by warping ${\mI}^{(t)}$ based on the predicted optical flow. A depth map can be predicted in the same way.

\subsubsection{Imagination}
Pixels at $t+1$ that are not visible at $t$ cannot be predicted by warping. So we augment warping-based prediction with an implicit prediction based on statistical regularities in scenes, 
generated by the imagination network $g_{\theta_\text{imag}}$: ${\mI'}_{\text{Imag}}^{(t+1)} = \sum_k {\mI'}_{k\text{Imag}}^{(t+1)} \odot {\boldsymbol\pi'}_{k\text{Imag}}^{(t+1)}$, and ${\mD'}_{\text{Imag}}^{(t+1)} = \sum_k {\mD'}_{k\text{Imag}}^{(t+1)} \odot {\boldsymbol\pi'}_{k\text{Imag}}^{(t+1)}$. This is similar to the way several other OCRL methods \citep{burgess2019monet,locatello2020object} reconstruct input image without using geometric knowledge, except that our "imagination" predicts the next image by additionally conditioning on the observer's motion and the estimated motion information of objects.

\subsubsection{Combining Warping-based Prediction and Imagination}
The final predicted images are pixel-wise weighted average of the prediction made by warping the current image and the corresponding prediction by Imagination network:
\begin{equation}\label{merge}
{\mI'}^{(t+1)} = {\mI'}_{\text{Warp}}^{(t+1)} \odot \mW_\text{Warp} + {\mI'}_{\text{Imag}}^{(t+1)} \odot (1 - \mW_\text{Warp})
\end{equation}
Here, an element of the weight map for the prediction based on warping $\mW_\text{Warp} \in \mathbb{R}^{w \times h}$ at $(p,q)$ is $W_\text{Warp}(p,q) = \max\{\sum_{k,i,j} w(i,j,p,q), 1\}$ where $w(i,j,p,q)$ is the the portion of pixel $(i,j)$ from $\mI^{(t)}$ that contribute to drawing pixel $(p,q)$ of ${\mI'}_{\text{Warp}}^{(t+1)}$. The intuition is that imagination is only needed when not enough pixels from $t$ will land near $(p,q)$ at $t+1$ based on the predicted optical flow. The same weighting applies for generating the final predicted depth ${\mD'}^{(t+1)}$.

\subsubsection{Learning Objective}
We have explained how to predict the spatial states of each object, ${\vx'}_{k}^{(t+1)}$ and ${p'}_{\phi_{k}}^{(t+1)}$, the next image ${\mI'}^{(t+1)}$ and depth map ${\mD'}^{(t+1)}$. Among the three prediction targets, only the ground truth for the next image is available as $\mI^{(t+1)}$ and therefore, the main learning objective is the prediction error for $\mI^{(t+1)}$ ($L_\text{image}=\text{MSE}({I'}^{(t+1)}, I^{(t+1)})$).

Additionally, we include two self-consistency losses between the predicted (based on $t-1$ and $t$) and inferred (based on $t+1$) object spatial states: the squared distance between predicted and inferred object location ( $ L_\text{location} =  \sum_{k=1}^{K} || \sum_{l=1}^{K+1} r_{kl}^{(t+1,t)} {\vx'}_{l}^{(t+1)} - \hat{\vx}_{k}^{(t+1)}||_2^2$) and the Jensen-Shannon divergence between predicted and inferred object pose ($ L_\text{pose} =  \sum_{k=1}^{K} D_\text{JS} (  \hat{\vp}_{\phi_{k}}^{(t+1)}|| \sum_{l=1}^{K+1} r_{kl}^{(t+1,t)} {\vp'}_{\phi_{l}}^{(t+1)})$, where $r_{kl}^{(t+1,t)}$ is the normalized soft-matching score between any extracted object $l$ in frames at $t$ and object $k$ at $t+1$ based on their identity code. 

Two regularization terms are introduced to avoid loss of gradient in training or trivial failure mode of learning. The term $L_\text{collapse} = - \min (\delta, ||\hat{\vx}_{k}- \hat{\vx}_{k,\text{batchshuffle}}||_1)$ penalizes too close distance between objects in one scene and a random scene in the same batch, which prevents a failure mode where the model assigns all objects' locations (across a dataset) to the same constant location. The term $L_\text{center} = \sum_{k=1}^{K} || \hat{\vx}_{k} - \sum_{i,j} \hat{\vm}_{i,j} \pi_{kij}  ||_2^2$ is the squared distance between the estimated 3D location of each object and average estimated 3D locations of all pixels weighted by their probability belonging to that object, which reflects a consistency assumption that points of an object should be around its center. The weights of these regularization terms and the self-consistency losses, are controlled via a single hyperparameter $\lambda$, manually tuned. All models are jointly trained to minimize the total loss:
\begin{equation}\label{loss}
L = L_\text{image} + \lambda (L_\text{location} + L_\text{pose} +  L_\text{center} +  L_\text{collapse})
\end{equation}

\subsection{Dataset}
We procedurally generated a dataset of triplets of images captured by a virtual camera with a field of view of 90$^{\circ}$ in a square room using Unity. The camera translates horizontally and pans with random small steps between consecutive frames to facilitate the learning of depth perception. 3 objects with random shape, size, surface color and textures are spawned at random locations in the room and each moves with a randomly selected constant velocity and panning speed. The translation and panning of the camera relative to its own reference frame and its intrinsic are known to the networks.
An important feature of this dataset is that more complex and diverse textures are used on both the objects and the background than other commonly used synthetic datasets for OCRL, such as CLEVRset \citep{johnson2017clevr} and GQN data \citep{eslami2018neural}. However, we also verify our model can generate coherent object masks from GQN data \citep{eslami2018neural} (ROOMS in \citep{henderson2020unsupervised}).

\section{Results}

\begin{table} 
    \begin{center}
    \begin{small}
    \begin{sc}
    \centering
    \begin{tabular}{@{}lll@{}} 
        \toprule
        Model & ARI-fg & IoU \\
        \midrule
        MONet-bigger & 0.36  & 0.20  \\
        slot-attention & 0.28$\pm$0.12(4)  & 0.27$\pm$0.08(4)  \\
        slot-attention-128 & 0.34  & 0.38 \\
        slate & 0.30 & 0.20 \\
        AMD & 0.19 & 0.02 \\
        O3V & 0.37$\pm$0.01(3) & 0.22$\pm$0.10(3) \\
        OPPLE (our model) & \textbf{0.58$\pm$0.07}(6) & \textbf{0.45$\pm$0.02}(6) \\
        \bottomrule
    \end{tabular}
    \end{sc}
    \end{small}
    \end{center}
     \vspace{-0.1in}
 \caption{Performance of models on object segmentation.}
 \label{table:segmentation}
 \vspace{-0.1in}
\end{table}

After training OPPLE and other representative or influential recent OCRL models on the dataset we generated, we evaluated their performance of object segmentation on 4000 test images unused during training but generated randomly with the same procedure. 
Some of these models learn from discrete images, while others also learn from videos but with limitations such as not inferring 3D information or not applicable to single images (O3V). Additionally, we demonstrated the abilities not attempted by most other models: inferring locations of objects in 3D space and the depth of the scenes.
We tested two versions of our model. The first makes with unnatural assumption that the equation for head motion-induced object apparent motion in 3D space and the equation for how pixels move with an object (rigid body motion) are known. The second learns both of these relationships with FC networks jointly with other parts of the model by the same loss function, relaxing these assumptions. We also confirmed our performance on GQN dataset.

\subsection{Object Segmentation}
Following prior works \citep{PMLRgreff19a, engelcke2019genesis,engelcke2021genesis}, we evaluated segmentation with the Adjusted Rand Index of foreground objects (ARI-fg). In addition, for each image, we matched ground-truth objects and background with each of the segmented class by ranking their Intersection over Union (IoU) and quantified the average IoU over entities. The performance is summarized in table \ref{table:segmentation}.\footnote{We trained MONet and slot-attention with both the original and increased network size. Note that O3V is evaluated on videos instead of images due to its requirement. We evaluated some models multiple times by training with different random seeds within our resource limit but could not do so for all models. When a model is trained multiple times, the performance is reported as mean $\pm$ standard deviation (number of experiments). } 

\begin{figure}
\begin{center}
\includegraphics[width=0.45\textwidth]{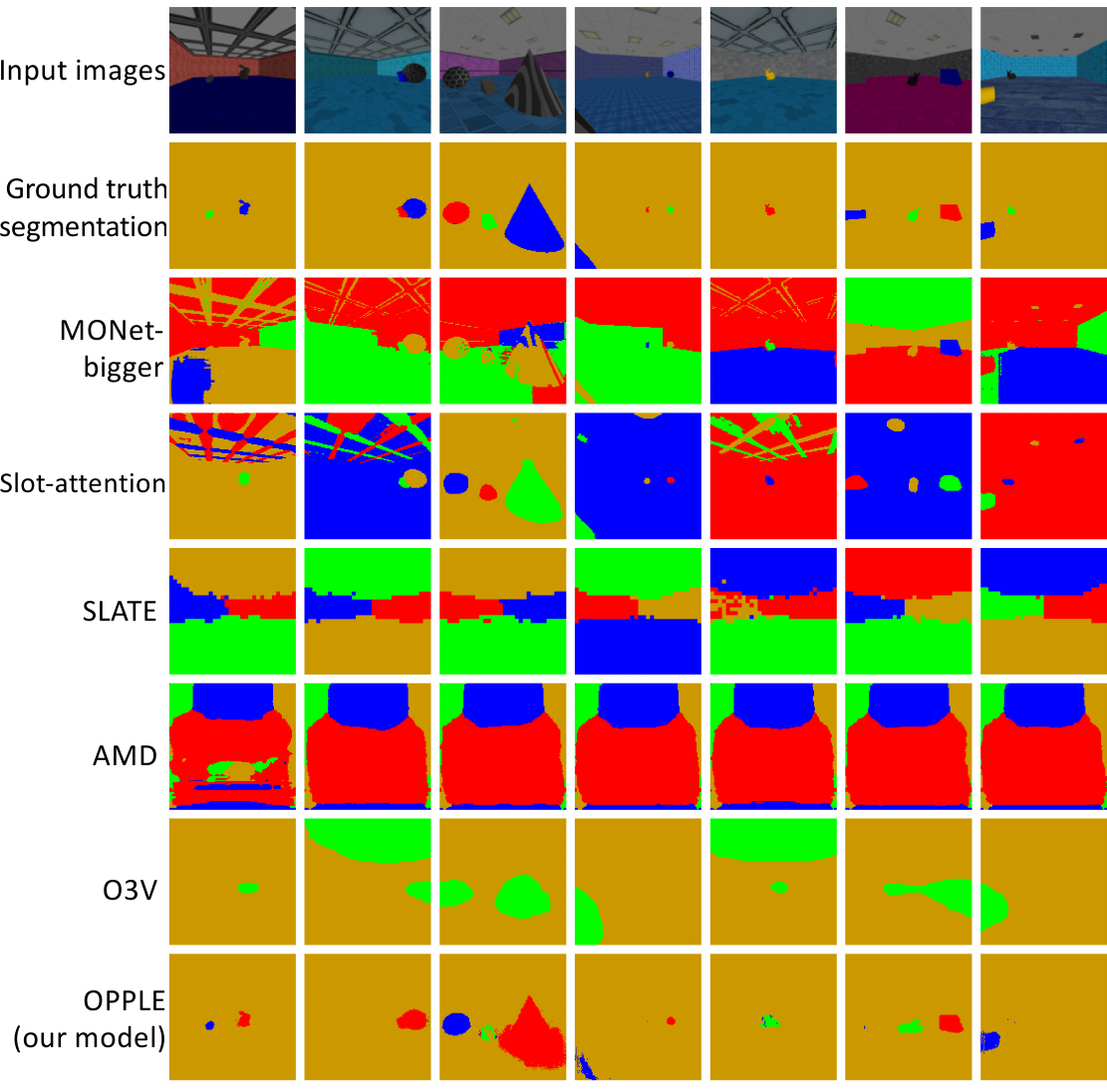}

\end{center}
\vspace{-0.2in}
\caption{Examples of object segmentation across models} 
\label{fig:segmentation}
\vspace{-0.1in}
\end{figure}

Our model outperforms all compared models on both metrics (Table \ref{table:segmentation}). 
As shown in Figure \ref{fig:segmentation}, MONet~\citep{burgess2019monet} appears to heavily rely on color to group pixels into the same masks and fails at complex texture. This may be because the model relies on a VAE bottleneck to compress data. Slot-attention~\citep{locatello2020object} tends to group distributed patches with similar colors or patterns in background as an object, which may reflect a strong reliance on texture similarity as a cue. SLATE~\citep{singh2021illiterate} appears to focus on segmenting regions of the background while ignoring objects. This may be because variation in colors/patterns/shades of backgrounds constitutes the major variation across images for the model to capture. We postulate there may be fundamental limitation in the approaches that learn purely from discrete images. 
The failure of AMD~\citep{liu2021emergence} is surprising since the model also learns from movement of objects. It may be because the translation of objects within the images are smaller than the border part of the scenes when the camera rotates, leading the model to focus on predicting the motion of pixels near the image borders (note that AMD predicts a single 2D motion vector for all pixels within a mask while OPPLE predicts 3D motion for each pixel). O3V~\citep{henderson2020unsupervised} tends to group multiple objects or adding ceiling as an object, perhaps because it requires larger camera or object motion. OPPLE is able to learn object-specific masks because they are used to predict optical flow for each object. A wrong segmentation would generate large prediction error at training. Such prediction error forces the network to learn features that identify object boundaries. We also evaluate our model on the GQN dataset with resulting performance of ARI-fg=0.37 and IoU=0.34.\footnote{Note that this dataset as used in previous work has an unnatural setting that camera moves on a ring, pointing to the center of a room. Example outputs are illustrated in supplementary material}

In order to determine the necessity of the warping module, we trained a version without it which resulted in much lower segmentation performance (ARI-fg=0.23  IoU=0.23) than the original model, indicating it's necessity.
We also found that ablating our regularizers (by setting $\lambda$ to $0$) decreases segmentation performance (ARI-fg=0.50, IoU=0.40). 

\vspace{-0.1in}

\begin{figure*}[h]
\begin{center}
\includegraphics[width=0.7\textwidth]{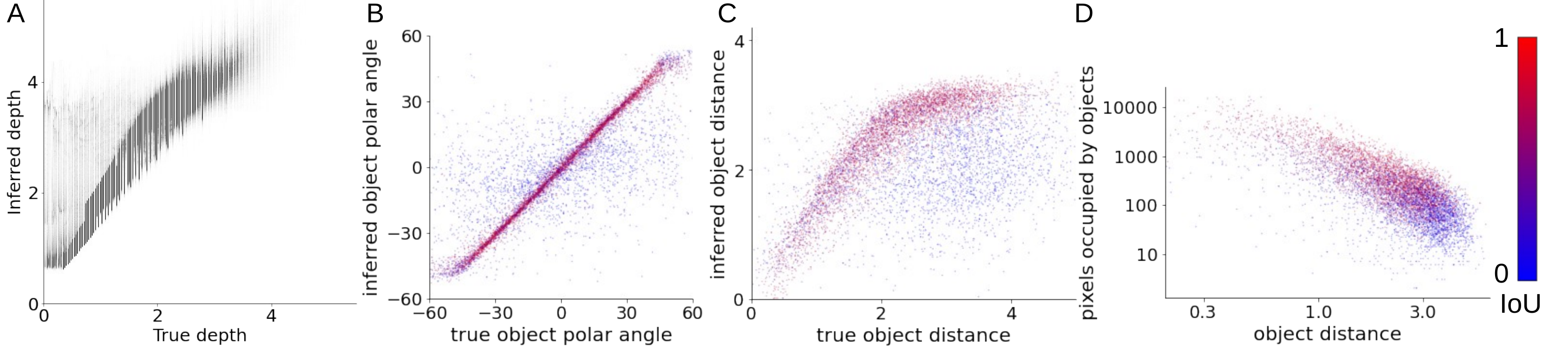}
\end{center}
\vspace{-0.2in}
\caption{Performance of depth inference (A) and 3D object localization (B,C). D: Nearer and larger objects are segmented better.} 
\label{fig:scatter_plot}
\vspace{-0.1in}
\end{figure*}
\subsection{Object Localization} 
The Object Extraction Network infers egocentric 3D object locations. We convert the inferred locations to viewing angles (from central line) and distance in polar coordinate relative to the camera. Figure \ref{fig:scatter_plot}B,C plot the true and inferred viewing angles and distance, color coded by objects' IoUs. Object viewing angles are better estimated (correlation $r=0.86$) than distances ($r=0.51$). Both viewing angles and distance of better segmented objects (red dots, colors indicating IoU) are estimated with higher accuracy (concentrated along the diagonal in B,C). Distance estimation is negatively biased for farther objects, potentially because the regularization term on the distance between the predicted and inferred object location at frame $t+1$ favors shorter distance when estimation is noisy. Because our objects did not move vertically, there is no teaching signal for the network to learn correct heights for object centers.

Generally, objects that are larger and closer to the camera are segmented better, likely because they exhibit larger optical flow in training (red dots in Figure \ref{fig:scatter_plot}D).

\subsection{Depth Perception}
OPPLE also learns to infer depth (distance of pixels from the camera). Figure \ref{fig:scatter_plot}A shows that the inferred depth are highly correlated with ground truth depth (correlation $r=0.92$). Because background occurs in every training sample, the network appears to bias the depth estimation on objects towards the depth of the walls behind, as shown by a dim cloud of points above diagonal line in the scatter plot.
We observed during training that the network first learns to infer global depth structure of the background before being able to infer objects' depth and segmentation.

\subsection{Relaxing Assumptions}
We trained and tested a version of our network in which the rules of rigid body motion and self-motion induced apparent motion are replaced by neural networks with two FC layers. On our custom dataset it obtained segmentation performance (ARI=0.58, IoU=0.44) comparable to the original model. However, estimation of object viewing angle (correlation $r=0.62$), object distance ($r=-0.01$), and depth ($r=0.33$) decreased. Further details are discussed in the Supplementary.

\section{Discussion}
We demonstrate that three capacities achieved by infants for object and depth perception can be learned without supervision using prediction as the main learning objective, while incorporating an assumption of rigidity of object that is also honored by early infants \citep{spelke1990principles}. Ablation studies suggest that encouraging consistency between the predicted and inferred latent object spatial information helps improve learning performance. Our work is at the cross point of two fields: object-centric representation~\citep{locatello2020object} and self-supervised learning~\citep{PMLRchen20j}. Many existing works in self-supervised learning do not produce object-based representation, 
but instead encode the entire scene as one vector or learn a feature map useful for downstream tasks~\citep{oquab2023dinov2}. 
Other OCRL models overcome this by assigning one representation to each object, as we do. Although works such as MONet~\citep{burgess2019monet}, IODINE~\citep{PMLRgreff19a}, slot-attention~\citep{locatello2020object}, GENESIS ~\citep{engelcke2019genesis} and PSGNet \citep{bear2020learning} can also segment objects and capture occlusion effect in images, few achieve all three capacities that we target purely by self-supervised learning, with the exception of a closely related work O3V ~\citep{henderson2020unsupervised} and a few others that learn 3D representation by rendering \citep{chen2021roots} \citep{crawford2020learning} \citep{stelzner2021decomposing}. One major distinction is that O3V and ROOTS need to process multiple frames to infer objects and require ground-truth camera locations relative to a global reference frame, while OPPLE can be applied to single images after training, and only needs information of egomotion relative to the observers during training.
Another distinction of OPPLE from \citep{henderson2020unsupervised} and many other works is that OPPLE learns from prediction instead of reconstruction. 
SQAIR \citep{kosiorek2018sequential} and SCALOR\citep{jiang2019scalor} represent another related direction for learning object representation in videos. A distinction is that they focus only on 2D motions within a frame. Interestingly, our ablation study shows that when the rigidity assumption is relaxed (by approximating rigid body motion rule with neural networks), depth perception and 3D localization both degrade. C-SWM~\citep{kipf2019contrastive} also learns object masks and predicts their future states utilizing Graphic Neural Network (GNN). The order of mapping object slots to nodes of GNN is fixed through time, which requires a fixed order of extracting objects. We solve this issue by a soft matching between object representations extracted from different time points. Although we use LSTM to sequentially extract multiple objects from single frames, this is not a critical choice and we envision it can be replaced by mechanism similar to slot attention. The major factor distinguishing OPPLE from slot-attention based models such as SAVi\citep{kipf2021conditional} and SAVi++\citep{elsayed2022savi++} is that they require external information of optical flow or depth which is unnatural to the brain (the brain of newborns are not equiped with these abilities), while our model can simultaneously learn to infer depth and predict optical flow together with object perception. 

 Our result also provides insights for brain research. In developmental psychology literature\citep{spelke1990principles}, a few more principles have been proposed to be honored by young infants for object perception, in addition to \textit{rigidity}: \textit{cohesion} (two surface points on the same object should be linked by a connected path of surface points and move continuously in time), \textit{boundedness} (two objects cannot occupy the same place at the same time), and \textit{no action at a distance} (independent motions indicates distinct objects). We note that \textit{boundedness} is essentially demanded by the formulation of probabilistic segmentation maps as in almost all segmentation model, and the assumption that one object is associated with one velocity essentially reflects \textit{no action at a distance}. The fact that OPPLE learns object perception without enforcing \textit{cohesion} (although we assumed objects have inertia) suggests it may not be as critical as other principles. However, it may be required in more complex environments than the ones we have tested on or may help improve the learning performance.

\bibliographystyle{named}
\bibliography{ijcai24}

\begin{thebibliography}{}

\bibitem[\protect\citeauthoryear{Bear \bgroup \em et al.\egroup }{2020}]{bear2020learning}
Daniel~M Bear, Chaofei Fan, Damian Mrowca, Yunzhu Li, Seth Alter, Aran Nayebi, Jeremy Schwartz, Li~Fei-Fei, Jiajun Wu, Joshua~B Tenenbaum, et~al.
\newblock Learning physical graph representations from visual scenes.
\newblock {\em arXiv preprint arXiv:2006.12373}, 2020.

\bibitem[\protect\citeauthoryear{Burgess \bgroup \em et al.\egroup }{2019}]{burgess2019monet}
Christopher~P Burgess, Loic Matthey, Nicholas Watters, Rishabh Kabra, Irina Higgins, Matt Botvinick, and Alexander Lerchner.
\newblock Monet: Unsupervised scene decomposition and representation.
\newblock {\em arXiv preprint arXiv:1901.11390}, 2019.

\bibitem[\protect\citeauthoryear{Chen \bgroup \em et al.\egroup }{2017}]{chen2017deeplab}
Liang-Chieh Chen, George Papandreou, Iasonas Kokkinos, Kevin Murphy, and Alan~L Yuille.
\newblock Deeplab: Semantic image segmentation with deep convolutional nets, atrous convolution, and fully connected crfs.
\newblock {\em IEEE transactions on pattern analysis and machine intelligence}, 40(4):834--848, 2017.

\bibitem[\protect\citeauthoryear{Chen \bgroup \em et al.\egroup }{2020}]{PMLRchen20j}
Ting Chen, Simon Kornblith, Mohammad Norouzi, and Geoffrey Hinton.
\newblock A simple framework for contrastive learning of visual representations.
\newblock In Hal~Daumé III and Aarti Singh, editors, {\em Proceedings of the 37th International Conference on Machine Learning}, volume 119 of {\em Proceedings of Machine Learning Research}, pages 1597--1607. PMLR, 13--18 Jul 2020.

\bibitem[\protect\citeauthoryear{Chen \bgroup \em et al.\egroup }{2021}]{chen2021roots}
Chang Chen, Fei Deng, and Sungjin Ahn.
\newblock Roots: Object-centric representation and rendering of 3d scenes.
\newblock {\em J. Mach. Learn. Res.}, 22:259--1, 2021.

\bibitem[\protect\citeauthoryear{Crawford and Pineau}{2020}]{crawford2020learning}
Eric Crawford and Joelle Pineau.
\newblock Learning 3d object-oriented world models from unlabeled videos.
\newblock In {\em Workshop on Object-Oriented Learning at ICML}, 2020.

\bibitem[\protect\citeauthoryear{Deng \bgroup \em et al.\egroup }{2009}]{deng2009imagenet}
Jia Deng, Wei Dong, Richard Socher, Li-Jia Li, Kai Li, and Li~Fei-Fei.
\newblock Imagenet: A large-scale hierarchical image database.
\newblock In {\em 2009 IEEE conference on computer vision and pattern recognition}, pages 248--255. Ieee, 2009.

\bibitem[\protect\citeauthoryear{Elsayed \bgroup \em et al.\egroup }{2022}]{elsayed2022savi++}
Gamaleldin~F Elsayed, Aravindh Mahendran, Sjoerd van Steenkiste, Klaus Greff, Michael~C Mozer, and Thomas Kipf.
\newblock Savi++: Towards end-to-end object-centric learning from real-world videos.
\newblock {\em arXiv preprint arXiv:2206.07764}, 2022.

\bibitem[\protect\citeauthoryear{Engelcke \bgroup \em et al.\egroup }{2019}]{engelcke2019genesis}
Martin Engelcke, Adam~R Kosiorek, Oiwi~Parker Jones, and Ingmar Posner.
\newblock Genesis: Generative scene inference and sampling with object-centric latent representations.
\newblock {\em arXiv preprint arXiv:1907.13052}, 2019.

\bibitem[\protect\citeauthoryear{Engelcke \bgroup \em et al.\egroup }{2021}]{engelcke2021genesis}
Martin Engelcke, Oiwi~Parker Jones, and Ingmar Posner.
\newblock Genesis-v2: Inferring unordered object representations without iterative refinement.
\newblock {\em arXiv preprint arXiv:2104.09958}, 2021.

\bibitem[\protect\citeauthoryear{Eslami \bgroup \em et al.\egroup }{2018}]{eslami2018neural}
SM~Ali Eslami, Danilo Jimenez~Rezende, Frederic Besse, Fabio Viola, Ari~S Morcos, Marta Garnelo, Avraham Ruderman, Andrei~A Rusu, Ivo Danihelka, Karol Gregor, et~al.
\newblock Neural scene representation and rendering.
\newblock {\em Science}, 360(6394):1204--1210, 2018.

\bibitem[\protect\citeauthoryear{Feinberg}{1978}]{feinberg1978efference}
Irwin Feinberg.
\newblock Efference copy and corollary discharge: implications for thinking and its disorders.
\newblock {\em Schizophrenia bulletin}, 4(4):636, 1978.

\bibitem[\protect\citeauthoryear{Greff \bgroup \em et al.\egroup }{2019}]{PMLRgreff19a}
Klaus Greff, Rapha{\"e}l~Lopez Kaufman, Rishabh Kabra, Nick Watters, Christopher Burgess, Daniel Zoran, Loic Matthey, Matthew Botvinick, and Alexander Lerchner.
\newblock Multi-object representation learning with iterative variational inference.
\newblock In Kamalika Chaudhuri and Ruslan Salakhutdinov, editors, {\em Proceedings of the 36th International Conference on Machine Learning}, volume~97 of {\em Proceedings of Machine Learning Research}, pages 2424--2433. PMLR, 09--15 Jun 2019.

\bibitem[\protect\citeauthoryear{Henderson and Lampert}{2020}]{henderson2020unsupervised}
Paul Henderson and Christoph~H Lampert.
\newblock Unsupervised object-centric video generation and decomposition in 3d.
\newblock {\em Advances in Neural Information Processing Systems}, 33:3106--3117, 2020.

\bibitem[\protect\citeauthoryear{Jiang \bgroup \em et al.\egroup }{2019}]{jiang2019scalor}
Jindong Jiang, Sepehr Janghorbani, Gerard De~Melo, and Sungjin Ahn.
\newblock Scalor: Generative world models with scalable object representations.
\newblock {\em arXiv preprint arXiv:1910.02384}, 2019.

\bibitem[\protect\citeauthoryear{Johnson \bgroup \em et al.\egroup }{2017}]{johnson2017clevr}
Justin Johnson, Bharath Hariharan, Laurens Van Der~Maaten, Li~Fei-Fei, C~Lawrence~Zitnick, and Ross Girshick.
\newblock Clevr: A diagnostic dataset for compositional language and elementary visual reasoning.
\newblock In {\em Proceedings of the IEEE conference on computer vision and pattern recognition}, pages 2901--2910, 2017.

\bibitem[\protect\citeauthoryear{Kipf \bgroup \em et al.\egroup }{2019}]{kipf2019contrastive}
Thomas Kipf, Elise van~der Pol, and Max Welling.
\newblock Contrastive learning of structured world models.
\newblock {\em arXiv preprint arXiv:1911.12247}, 2019.

\bibitem[\protect\citeauthoryear{Kipf \bgroup \em et al.\egroup }{2021}]{kipf2021conditional}
Thomas Kipf, Gamaleldin~F Elsayed, Aravindh Mahendran, Austin Stone, Sara Sabour, Georg Heigold, Rico Jonschkowski, Alexey Dosovitskiy, and Klaus Greff.
\newblock Conditional object-centric learning from video.
\newblock {\em arXiv preprint arXiv:2111.12594}, 2021.

\bibitem[\protect\citeauthoryear{Kosiorek \bgroup \em et al.\egroup }{2018}]{kosiorek2018sequential}
Adam Kosiorek, Hyunjik Kim, Yee~Whye Teh, and Ingmar Posner.
\newblock Sequential attend, infer, repeat: Generative modelling of moving objects.
\newblock {\em Advances in Neural Information Processing Systems}, 31, 2018.

\bibitem[\protect\citeauthoryear{Li \bgroup \em et al.\egroup }{2020}]{li2020learning}
Nanbo Li, Cian Eastwood, and Robert Fisher.
\newblock Learning object-centric representations of multi-object scenes from multiple views.
\newblock {\em Advances in Neural Information Processing Systems}, 33:5656--5666, 2020.

\bibitem[\protect\citeauthoryear{Liu \bgroup \em et al.\egroup }{2021}]{liu2021emergence}
Runtao Liu, Zhirong Wu, Stella Yu, and Stephen Lin.
\newblock The emergence of objectness: Learning zero-shot segmentation from videos.
\newblock {\em Advances in Neural Information Processing Systems}, 34:13137--13152, 2021.

\bibitem[\protect\citeauthoryear{Locatello \bgroup \em et al.\egroup }{2020}]{locatello2020object}
Francesco Locatello, Dirk Weissenborn, Thomas Unterthiner, Aravindh Mahendran, Georg Heigold, Jakob Uszkoreit, Alexey Dosovitskiy, and Thomas Kipf.
\newblock Object-centric learning with slot attention.
\newblock {\em arXiv preprint arXiv:2006.15055}, 2020.

\bibitem[\protect\citeauthoryear{Oquab \bgroup \em et al.\egroup }{2023}]{oquab2023dinov2}
Maxime Oquab, Timothée Darcet, Théo Moutakanni, Huy Vo, Marc Szafraniec, Vasil Khalidov, Pierre Fernandez, Daniel Haziza, Francisco Massa, Alaaeldin El-Nouby, Mahmoud Assran, Nicolas Ballas, Wojciech Galuba, Russell Howes, Po-Yao Huang, Shang-Wen Li, Ishan Misra, Michael Rabbat, Vasu Sharma, Gabriel Synnaeve, Hu~Xu, Hervé Jegou, Julien Mairal, Patrick Labatut, Armand Joulin, and Piotr Bojanowski.
\newblock Dinov2: Learning robust visual features without supervision, 2023.

\bibitem[\protect\citeauthoryear{O'Reilly \bgroup \em et al.\egroup }{2017}]{o2017deep}
Randall~C O'Reilly, Dean~R Wyatte, and John Rohrlich.
\newblock Deep predictive learning: a comprehensive model of three visual streams.
\newblock {\em arXiv preprint arXiv:1709.04654}, 2017.

\bibitem[\protect\citeauthoryear{Ronneberger \bgroup \em et al.\egroup }{2015}]{ronneberger2015u}
Olaf Ronneberger, Philipp Fischer, and Thomas Brox.
\newblock U-net: Convolutional networks for biomedical image segmentation.
\newblock In {\em International Conference on Medical image computing and computer-assisted intervention}, pages 234--241. Springer, 2015.

\bibitem[\protect\citeauthoryear{Seitzer \bgroup \em et al.\egroup }{2022}]{seitzer2022bridging}
Maximilian Seitzer, Max Horn, Andrii Zadaianchuk, Dominik Zietlow, Tianjun Xiao, Carl-Johann Simon-Gabriel, Tong He, Zheng Zhang, Bernhard Sch{\"o}lkopf, Thomas Brox, et~al.
\newblock Bridging the gap to real-world object-centric learning.
\newblock {\em arXiv preprint arXiv:2209.14860}, 2022.

\bibitem[\protect\citeauthoryear{Singh \bgroup \em et al.\egroup }{2021}]{singh2021illiterate}
Gautam Singh, Fei Deng, and Sungjin Ahn.
\newblock Illiterate dall-e learns to compose.
\newblock In {\em International Conference on Learning Representations}, 2021.

\bibitem[\protect\citeauthoryear{Spelke}{1990}]{spelke1990principles}
Elizabeth~S Spelke.
\newblock Principles of object perception.
\newblock {\em Cognitive science}, 14(1):29--56, 1990.

\bibitem[\protect\citeauthoryear{Stelzner \bgroup \em et al.\egroup }{2021}]{stelzner2021decomposing}
Karl Stelzner, Kristian Kersting, and Adam~R Kosiorek.
\newblock Decomposing 3d scenes into objects via unsupervised volume segmentation.
\newblock {\em arXiv preprint arXiv:2104.01148}, 2021.

\end{thebibliography}


\begin{thebibliography}{}

\bibitem[\protect\citeauthoryear{Burgess \bgroup \em et al.\egroup }{2019}]{burgess2019monet}
Christopher~P Burgess, Loic Matthey, Nicholas Watters, Rishabh Kabra, Irina Higgins, Matt Botvinick, and Alexander Lerchner.
\newblock Monet: Unsupervised scene decomposition and representation.
\newblock {\em arXiv preprint arXiv:1901.11390}, 2019.

\bibitem[\protect\citeauthoryear{Elsayed \bgroup \em et al.\egroup }{2022}]{elsayed2022savi++}
Gamaleldin~F Elsayed, Aravindh Mahendran, Sjoerd van Steenkiste, Klaus Greff, Michael~C Mozer, and Thomas Kipf.
\newblock Savi++: Towards end-to-end object-centric learning from real-world videos.
\newblock {\em arXiv preprint arXiv:2206.07764}, 2022.

\bibitem[\protect\citeauthoryear{Eslami \bgroup \em et al.\egroup }{2018}]{eslami2018neural}
SM~Ali Eslami, Danilo Jimenez~Rezende, Frederic Besse, Fabio Viola, Ari~S Morcos, Marta Garnelo, Avraham Ruderman, Andrei~A Rusu, Ivo Danihelka, Karol Gregor, et~al.
\newblock Neural scene representation and rendering.
\newblock {\em Science}, 360(6394):1204--1210, 2018.

\bibitem[\protect\citeauthoryear{Greff \bgroup \em et al.\egroup }{2022}]{greff2022kubric}
Klaus Greff, Francois Belletti, Lucas Beyer, Carl Doersch, Yilun Du, Daniel Duckworth, David~J Fleet, Dan Gnanapragasam, Florian Golemo, Charles Herrmann, et~al.
\newblock Kubric: A scalable dataset generator.
\newblock In {\em Proceedings of the IEEE/CVF Conference on Computer Vision and Pattern Recognition}, pages 3749--3761, 2022.

\bibitem[\protect\citeauthoryear{Henderson and Lampert}{2020}]{henderson2020unsupervised}
Paul Henderson and Christoph~H Lampert.
\newblock Unsupervised object-centric video generation and decomposition in 3d.
\newblock {\em Advances in Neural Information Processing Systems}, 33:3106--3117, 2020.

\bibitem[\protect\citeauthoryear{Kingma and Ba}{2014}]{kingma2014adam}
Diederik~P Kingma and Jimmy Ba.
\newblock Adam: A method for stochastic optimization.
\newblock {\em arXiv preprint arXiv:1412.6980}, 2014.

\bibitem[\protect\citeauthoryear{Locatello \bgroup \em et al.\egroup }{2020}]{locatello2020object}
Francesco Locatello, Dirk Weissenborn, Thomas Unterthiner, Aravindh Mahendran, Georg Heigold, Jakob Uszkoreit, Alexey Dosovitskiy, and Thomas Kipf.
\newblock Object-centric learning with slot attention.
\newblock {\em arXiv preprint arXiv:2006.15055}, 2020.

\bibitem[\protect\citeauthoryear{Pouget \bgroup \em et al.\egroup }{2000}]{pouget2000information}
Alexandre Pouget, Peter Dayan, and Richard Zemel.
\newblock Information processing with population codes.
\newblock {\em Nature Reviews Neuroscience}, 1(2):125--132, 2000.

\end{thebibliography}

\end{document}


\renewcommand{\thefigure}{S\arabic{figure}}
\setcounter{figure}{0}

\maketitle


\section{Additional demonstration of OPPLE's inference and prediction}
\subsection{Visualization of outputs by the learned networks and prediction for the next frame}
\begin{figure}[h]
\begin{center}
\includegraphics[width=1.0\textwidth]{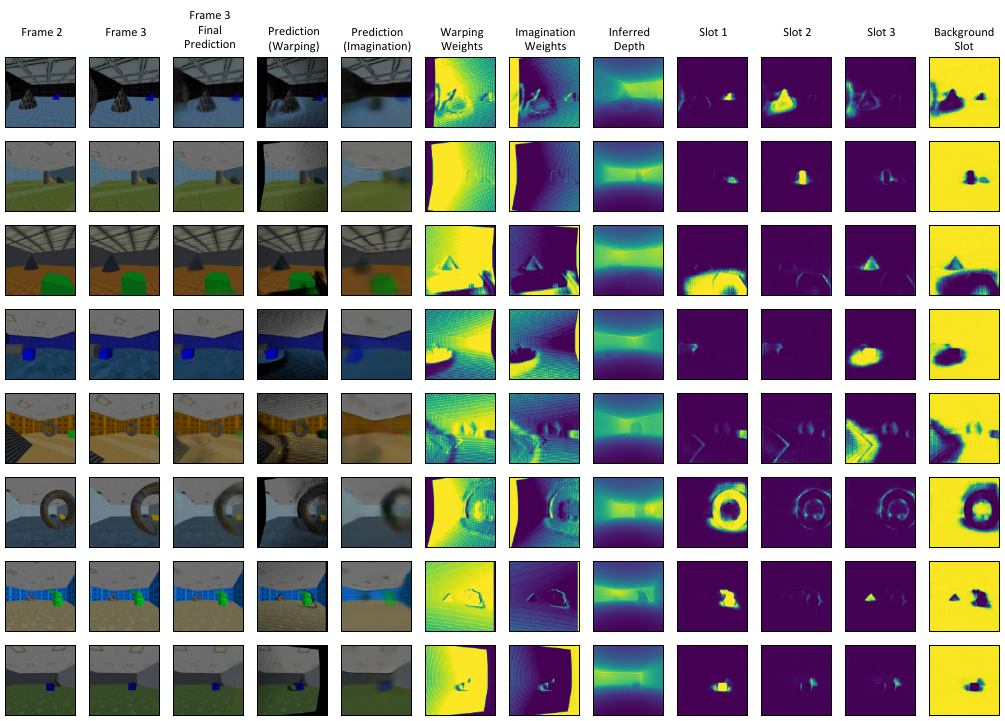}
\end{center}
\vspace{-0.1in}
\caption{OPPLE's internal segmentation, depth inference and prediction for triplets in our dataset} 
\label{fig:supp1}
\vspace{-0.1in}
\end{figure}

As visualized in Figure \ref{fig:supp1}, Frame 2 (column 1) is the basis of prediction (frame 1 is not shown). Frame 3 (ground truth in column 2) is predicted (column 3) by combining two approaches: warping-based prediction (column 4) and imagination (column 5). The warping-based prediction is made based on the optical flow predicted from the inferred depth (column 8), object segmentation (the last 4 columns), inferred object location and velocity, and the camera's motion. In this experiment, the maximum possible number of visible objects is set to 3.

As expected it can be seen that warping-based prediction generates sharper images (column 4) than imagination-based prediction (column 5). This is because the latter is based on learned statistical regularity of environments and is only needed for small regions in the image (thus receiving less teaching signal).
This relaxation of the need to encode fine details of a scene in the imagination network is advantageous because the goal of object perception is primarily to obtain an estimation of 3D information and coarse features of an object. The warping-based prediction also does not require encoding fine details of each object.
The warping weights (column 6) is the total contribution of warping-based prediction towards the final prediction in each pixel.

\subsection{Depth perception}
We demonstrate a few example images, their ground truth depth and the depth inferred by OPPLE. Our network can capture the global 3D structure of the scene, although the estimation of the depth of the objects tends to be biased towards the depth of the background behind them, consistent with the scatter plot in the main paper. We suspect that by introducing more translational motion of the camera parallel to the image plane will allow the depth network to learn more accurate depth inference for objects, as such camera movement induces strong motion parallax effect.
We see evidence for this in our results on the GQN dataset, as discussed in the next section.

We would like to emphasize that the depth inference is performed on single images without the need of motion cues, unlike other unsupervised models such as O3V and Savi++ that need videos at both training and test time to learn and infer 3D information.

\begin{figure}[h]
\begin{center}
\includegraphics[width=1.0\textwidth]{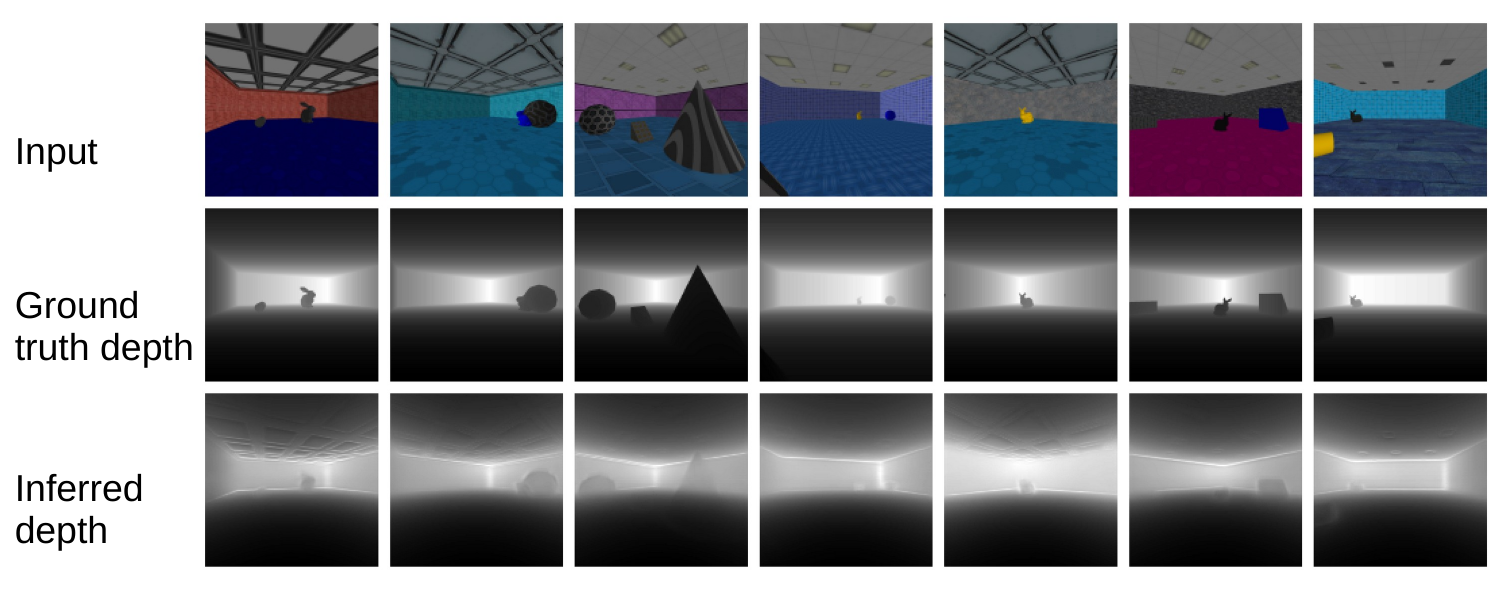}
\end{center}
\vspace{-0.2in}
\caption{Comparison between ground truth depth and inferred depth} 
\label{fig:supp2}
\vspace{-0.1in}
\end{figure}

\subsection{Performance on GQN dataset}\label{appendix:GQN}
We evaluated our model on the ROOMS dataset using the rendering code provided by \citep{henderson2020unsupervised}. The dataset is based on the GQN rooms-ring-camera dataset \citep{eslami2018neural} but we allowed all motion speeds to be uniformly sampled from an interval instead of being chosen from narrow Gaussian distributions as in  \citep{henderson2020unsupervised}. 
 It appears that the depth images (column 8) are sharper and exhibit less bias for objects' depth towards longer distance than in Figure \ref{fig:supp1} (on our custom dataset). This is potentially because in this dataset, the dominant motion of the camera is parallel to the image plane (along the ring while looking to a direction close to the center of the room), consistent with our hypothesis above. This visualization also demonstrates that although we set a maximum number of slots (objects) visible to the camera (3 in these experiments), the network is able to output empty object masks when the actual number of visible objects is smaller than the number of slots. In other words, it is not required for the network to know how many objects are in the scene. Only a maximum number of slots needs to be set for the LSTM within the object extraction network.

\begin{figure} 
\begin{center}
\includegraphics[width=1.0\textwidth]{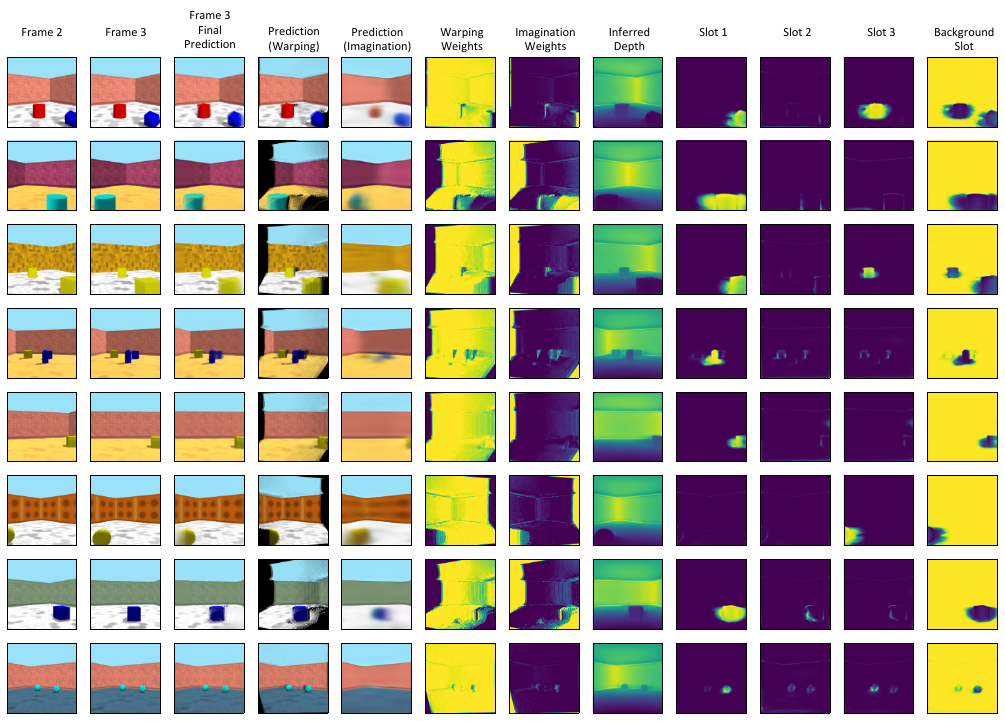}
\end{center}
\vspace{-0.2in}
\caption{Visualization of the two approaches of prediction (warping and imagination), their combination weights, inferred depth, and probabilistic segmentation maps from OPPLE on scenes of moving objects in a simulated room based on the ring environment of GQN\citep{eslami2018neural}, based on code from \citep{henderson2020unsupervised}.} 
\label{fig:supp3}
\vspace{-0.1in}
\end{figure}

As shown below, the depth inference of OPPLE on GQN dataset reaches superior performance of $r$=0.99 between ground truth depth and inferred depth (excluding pixels in the sky, because the sky's distance is effectively infinity in the ground truth). A scatter plot of inferred vs. ground truth depth for pixels excluding the sky is shown below (the vertical stripes are likely due to rounding issue in saving depth map and down-sampling).

\begin{figure} [H]
\begin{center}
\includegraphics[width=0.5\textwidth]{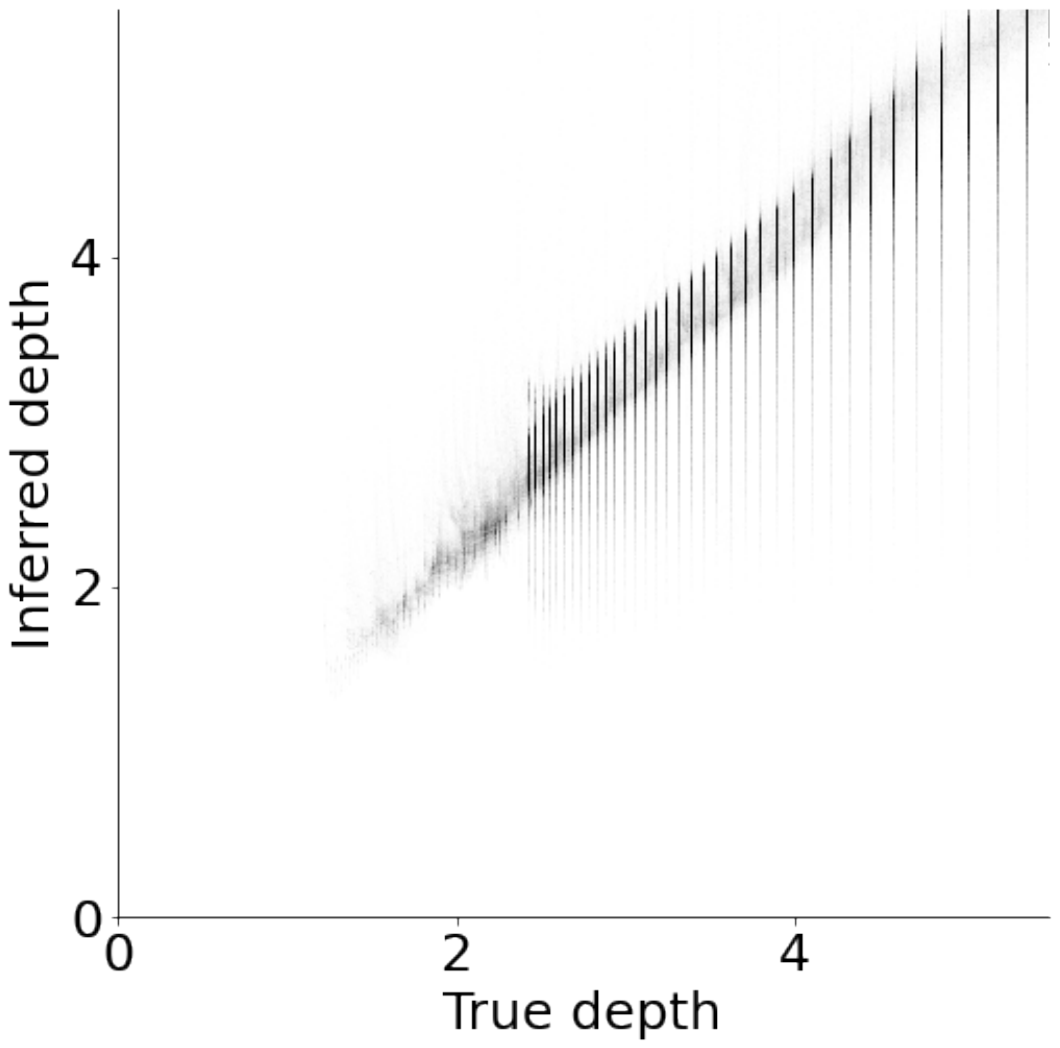}
\end{center}
\vspace{-0.7in}
\caption{Performance of depth inference of OPPLE on GQN dataset.} 
\label{fig:supp4}
\vspace{-0.1in}
\end{figure}

\section{Pseudo code of the OPPLE framework} \label{pseudocode}

\begin{algorithm}[H]
\caption{Object Perception by Predictive LEarning (OPPLE)}\label{alg:POP}
\begin{algorithmic}
\State \text{initialize} $f_{\theta_\text{obj}}$, $h_{\theta_\text{depth}}$, $g_{\theta_\text{imag}}$
\Require images $\mI^{(t-1)}, \mI^{(t)}, \mI^{(t+1)} \in \sR^{w \times h \times 3}$, self-motion $\vv_\text{obs}^{(t-1)},  \omega_\text{obs}^{(t-1)}, \vv_\text{obs}^{(t)}, \omega_\text{obs}^{(t)}$

\Ensure  prediction $\mI'^{(t+1)}$, segmentation $\boldsymbol\pi_{1:K+1}^{(t-1)}$, $\boldsymbol\pi_{1:K+1}^{(t)}$,  objects' codes  $\vz_{1:K}^{(t-1)}$, $\vz_{1:K}^{(t)}$, objects' locations and poses $\hat{\vx}_{1:K}^{(t-1)}, \vp_{\phi_{1:K}}^{(t-1)}, \hat{\vx}_{1:K}^{(t)}, \vp_{\phi_{1:K}}^{(t)}$
\For {$\tau = \{t-1 , t\}$ }
\State scene code $ e^{(\tau)} \gets \text{U-NetEncoder}_{f_{\theta_\text{obj}}}(\mI^{(\tau)})$
\State {$\text{object code }  \vz_{1:K}^{(\tau)}, \text{location } \hat{\vx}_{1:K}^{(\tau)}, \text{pose } \vp_{\phi_{1:K}}^{(\tau)} \gets \text{LSTM}_{f_{\theta_\text{obj}}}(e^{(\tau)}) $}
\State $\text{background code } \vz_{K+1}=0$
\State $\text{depth } \mD^{(\tau)} \gets h_{\theta_\text{depth}} (\mI^{(\tau)})$
\State {$\text{segmentation mask } \boldsymbol\pi_{1:K+1}^{(\tau)} \gets \text{Softmax ([U-NetDecoder}_{f_{\theta_\text{obj}}}(\mI^{(\tau)}, \vz_{1:K}^{(\tau)}), 0]) $ }

\EndFor
\State $\text{object matching scores } r_{kl} \gets \frac{\text{RBF}(\vz_{k}^{(t)}, \vz_{l}^{(t-1)})}{\sum_{m=1}^{K+1} \text{RBF}(\vz_{k}^{(t)}, \vz_{m}^{(t-1)})}, k, l \in 1:K+1 $
\For {$k \gets 1$ to $K$ }
\State $\text{object motion } \hat{\vv}_{1:K}, \boldsymbol\omega_{1:K} \gets r_{k,l}, \hat{\vx}_{k}^{(t)}, \hat{\vx}_{l}^{(t-1)}, \vp_{\phi_{k}}^{(t)}, \vp_{\phi_{l}}^{(t-1)}, \vv_\text{obs}^{(t-1)},  \omega_\text{obs}^{(t-1)}, l=1:K+1 $
\State $\text{object-specific optical flow}_k \gets \hat{\vv}_{1:K}, \boldsymbol\omega_{1:K}, \vv_\text{obs}^{(t)}, \omega_\text{obs}^{(t)}, \mD^{(t)},\hat{\vx}_{k}^{(t)} $
\EndFor
\State \text{warping-based prediction} ${\mI'}_\text{warp}^{(t+1)},  \mW_\text{Warp} \gets \text{Warp}(\mI^{(t)}, \text{optical flow}_{1:K+1}, \boldsymbol\pi_{1:K+1}^{(\tau)} )$  
\State \text{imagination} ${\mI'}_\text{imagine}^{(t+1)} \gets g_{\theta_\text{obj}}( \mI^{(t)} \odot \boldsymbol\pi_{1:K+1}^{(t)}, \log(\mD^{(t)}) \odot \boldsymbol\pi_{1:K+1}^{(t)},  \vv_\text{obs}, \omega_\text{obs}, \hat{\vv}_{1:K}, \hat{\vx}_{1:K})$  
\State final image prediction: ${\mI'}^{(t+1)} \gets {\mI'}_\text{warp}^{(t+1)}, {\mI'}_\text{imagine}^{(t+1)}, \mW_\text{Warp}$
\State update parameters: $\theta \gets \theta - \gamma \nabla_\theta [ |{\mI'}^{(t+1)} - {\mI}^{(t+1)}|^2 + \text{regularization loss} ] $, $\theta \in [{\theta_\text{obj}}, {\theta_\text{depth}}, {\theta_\text{imag}}]$ 
\end{algorithmic}
\end{algorithm}

\section{Network training and dataset} \label{implementation-details}

We trained the three networks jointly using ADAM optimization \cite{kingma2014adam} with a learning rate of $3\text{e}-4$, $\epsilon=1\text{e}-6$ and other default setting in PyTorch, with a batch size of 20. 55 epochs were trained on the dataset. We set $\lambda=1.0$.

The model was implemented in PyTorch and trained on NVidia RTX 6000 for about 8 days. We will release the dataset upon publication of the manuscript.

In consideration that other models originally demonstrated on simple datasets with homogeneous colors may need to increase their size for our dataset, we tested MONet\citep{burgess2019monet} and slot attention\citep{locatello2020object} with both the original network sizes and increased sizes. 
For MONet, we tested the original number of channels ([32, 32, 64, 64]) for the hidden layers of encoder of the component VAE (IoU=0.12, ARI-fg=0.27), and increased them to [32, 64, 128, 256, 256] with one more layer while also increasing the channel number in its attention network from 64 to 128 (MONet-bigger as reported in the main text: IoU=0.20, ARI-fg=0.36). The decoder layers' sizes were increased accordingly. 
For slot attention, we tested a variant which increased the number of features in the attention component from 64 to 128 (slot-attention-128, which took 15 days to train) in addition to the original setting (slot-attention) and found the performance to be similar. 
For our model, we tuned sizes of networks based on performance on the training set.

Images in the dataset are rendered in a custom Unity environment at a resolution of 512 $\times$ 512 and downsampled to 128 $\times$ 128 resolution for training. Images are rendered in sequence of 7 time steps for each scene. In each scene, a room with newly selected textures and objects is created.
The starting location and orientation of the camera and objects are initialized randomly.
The camera moves with random steps and pivots with random angles between consecutive frames; each object moves with random constant velocity drawn independently.
All possible valid sequential triplets out of the recorded 7 frames (e.g., frames 1,2,3; 1,3,5; 1,4,7, etc.) form our training samples, 
excluding the frame sets where objects collide with each other or the camera.
For our dataset we use 90k scenes for training which give us 275k valid triplets. 

\section{Details of prediction methods}

\subsection{Prediction of new location and pose of an object}\label{appendix:obj_pred}
We first explain how to use the inferred 3D spatial states of an object in two frame to estimate its velocity and predict its location in the next frame. Following the notation in 2.1 of the main text, in order to calculate the instantaneous velocity of object $k$ at $t-1$ from the perspective of the camera at $t$ (with a static reference frame), we need to first calculate the 3D location that object $k$ observed at $t-1$ should appear at $t$ to the camera if it is static relative to the background: $\hat{x}_k^{t-1 \rightarrow t} = \mM_{-\omega_\text{obs}}^{(t-1)} ( \hat{\vx}_{k}^{(t-1)}  - \vv_\text{obs}^{(t-1)} )$, where $\hat{\vx}_{k}^{(t-1)}$ is the inferred 3D location of object $k$ at $t-1$; $\vv_\text{obs}^{(t-1)}$ is the velocity of the camera from $t-1$ to $t$ viewed from a static reference frame centered at its original location; and $\mM_{-\omega_\text{obs}}^{(t-1)}$ is the rotation matrix $\begin{bmatrix}
\cos \omega_\text{obs}^{(t-1)} & \sin \omega_\text{obs}^{(t-1)} & 0\\
- \sin \omega_\text{obs}^{(t-1)} & \cos \omega_\text{obs}^{(t-1)} & 0 \\
0 & 0 & 1
\end{bmatrix}$ induced by the rotational speed of the camera from $t-1$ to $t$. Thus, the spatial offset from any object $j$ at $t-1$ to object $k$ at $t$, as viewed from the camera, is $\hat{\vv}_{j \rightarrow k}^{(t)} = \hat{x}_j^{(t)} - \hat{x}_j^{t-1 \rightarrow t}$. We calculate the velocity of object $k$ as the weighted average of the spatial offset against all candidate objects detected at $t-1$ using the matching scores $r_{kj}$ between object $k$ and those candidate objects: $\hat{\vv}_{k}^{(t)} = \sum_{j=0}^{K} r_{kj} \hat{\vv}_{j \rightarrow k}^{(t)}$

The spatial location where the object $k$ observed at $t$ would be at $t+1$ can be predicted as $ {\vx'}_k^{(t+1)} = \mM_{-\omega_\text{obs}}^{(t)} ( \hat{\vx}_{k}^{(t)} + \hat{\vv}_{k}^{(t)} - \vv_\text{obs}^{(t)} )$, with $\vv_\text{obs}^{(t)}$  and $\mM_{-\omega_\text{obs}}^{(t)}$ defined similarly as above, but for camera movement between $t$ and $t+1$.

As the matrix multiplication involved in this prediction is almost impossible to be hard-wired in the brain, we also tested the possibility of learning to approximate this computation with a fully-connected neural network, 
as is further discussed in section \ref{deep_warp} of this Supplementary.

As described in the main text, we use $\vp_{\phi_{k}}^{(t)} = [p_{\phi_{k,1}}^{(t)}, p_{\phi_{k,2}}^{(t)}, \cdots , p_{\phi_{k,b}}^{(t)}]$ to represent the probabilities that the pose of object $k$ at $t$ falls into $b$ equally spaced bins of angles in $(0, 2\pi)$, where $\sum_i p_{\phi_{k,i}}^{(t)} = 1$. With this notation, if we consider $b$ possible discrete rotational speeds $-\frac{(b-2)\pi}{b}, -\frac{(b-4)\pi}{b}, \cdots, 0, \cdots,  \frac{b\pi}{b}$, the probability that the rotational speed $\omega_{j \rightarrow k}^{(t)}$ is equal to any $\gamma_1$ among these discrete values if the same object is assigned index $j$ at $t-1$ and $k$ at $t$ is $p(\omega_{j \rightarrow k}^{(t)} = \gamma_1) \propto \sum_{l,m} p_{\phi_{j,l}}^{(t-1)} p_{\phi_{k,m}}^{(t-1)} \mathbbm{1}(\frac{(m-l)2\pi}{b}  \in \{\gamma_1 - 2\pi, \gamma_1, \gamma_1 + 2\pi\})$. We then calculate the likelihood of rotational speed for object $k$ similarly by weighting the likelihoods calculated for all possible candidate pairs $j,k$ with their matching scores. 

The probability that the pose of object $k$ is equal to any possible discrete yaw angle bin $\gamma_2$ at $t+1$ can be predicted through  ${p'}(\phi_k^{(t+1)} + \omega_{\text{obs}}^{(t)} = \gamma_2) = \sum\limits_{\substack{\gamma_1, \omega, \\ \gamma_2 - \gamma_1  \in  \{\omega-2\pi, \omega, \omega+2\pi \} }} p(\omega_{j \rightarrow k}^{(t)} = \omega) p( \phi_k^{(t)} = \gamma_1) $, where $\omega_{\text{obs}}^{(t)}$ is the angular velocity of the observer. Since we model $\phi_k^{(t)}$ and $\omega_k^{(t)}$ as probability distributions on discrete bins of angle while $\omega_{\text{obs}}$ can take continuous values, we convert ${p'}(\phi_k^{(t+1)} + \omega_{\text{obs}})$ to ${p'}(\phi_k^{(t+1)})$ on the same set of bins as $\phi_k^{(t)}$ by interpolation with a Von Mises kernel.

For the purpose of obtaining a point estimation of the rotational speed of $\omega_k$ to be used in predicting pixel-wise optical flow, we multiply the estimated probability for each angle bin with a vector pointing to the direction of that angle and take the angle of the sum of all these products (resembling the population vector code commonly observed in the brain for angular variables such as arm reaching direction or visual motion direction\citep{pouget2000information}).

\subsection{Camera intrinsic}\label{appendix:warping}
The Depth Perception network uses visual features in the image $\mI^{(t)}$ to infer the distance of all pixels to the camera (depth) $\mD^{(t)} \in \mathbb{R}^{w \times h} = h_{\theta}(\mI^{(t)})$. With the inferred depth $D^{(t)} (i,j)$ for a pixel at any coordinate $(i,j)$ in the image and the focal length $f$ of the camera, the pixel's 3D location as viewed by the camera can be determined as $\hat{\vm}_{(i,j)}^{(t)} =  {\frac {D^{(t)} (i,j)} {\sqrt{i^2+j^2+f^2}} \cdot [i,f,j]}$. Here, we take the pixel coordinate of the center of an image as (0,0) and pixel indices are assumed to be symmetric around the center of the image: $|i| \leq  \frac{w-1}{2}$,$|j| \leq  \frac{h-1}{2}$. Conversely, given a 3D coordinate location $x,y,z$ relative to the camera, the pixel coordinate it appears on the image can be calculated as well: $[i,j] = \frac{f}{y} [x,z]$. And its depth $d = \sqrt{x^2+y^2+z^2}$. This is the knowledge of camera intrinsic, which we assume is known. However, since these are relatively simple mapping functions, we think it is possible to learn to approximate them with neural networks together with other networks in future works.

\subsection{Weights in warping-based prediction and combination of two streams of predictions}\label{appendix:warping_weight}
The details of the procedure of predicting where a pixel at $t$ would land at $t+1$ on the image has been detailed in the main text. Here we describe the calculation of the weights when drawing the colors for every pixel at $t+1$.

As the object attribution of each pixel is not known but is inferred by $f_\text{obj}(\mI^{(t)})$, it is represented for every pixel as a probability of belonging to each object and the background $\boldsymbol{\pi}_{k}^{(t)}$, $k=1,2,\cdots,K+1$. Therefore, the predicted motion of each pixel should be described as a probability distribution over $K+1$ discrete target locations  $p({\vm'}_{(i,j)}^{(t+1)}) = \sum_{k=1}^{K+1} \pi_{kij}^{(t)} \cdot \delta({\vm'}_{k,(i,j)}^{(t+1)})$, i.e., pixel $(i,j)$ has a probability of $\pi_{kij}^{(t)}$ to move to location ${\vm'}_{k,(i,j)}^{(t+1)}$ at the next time point, for  $k=1,2,\cdots,K+1$.
With such probabilistic prediction of pixel movement for all visible pixel $(i,j)^{(t)}$, we can partially predict the colors of the next image at the pixel grids where some original pixels from the current view will land nearby by weighting their contribution:
\begin{equation}\label{eq:warping}
{\mI'}_{\text{Warp}}^{(t+1)}(p,q) = \begin{cases}
\frac {\sum_{k,i,j} w_k(i,j,p,q) I^{(t)}(i,j)} {\sum_{k,i,j} w_k(i,j,p,q)} , &\text{if } \sum_{k,i,j} w_k(i,j,p,q) > 0 \\
0, &\text{otherwise}
\end{cases}
\end{equation}
We define the weight of the contribution from any source pixel ${(i,j)}$ to a target pixel ${(p,q)}$ as 
\begin{equation}\label{warping_weight}
w_k(i,j,p,q) =   \pi_{kij}^{(t)} \cdot  e^{-\beta \cdot {{D'}_k^{(t+1)}(i,j)}} \cdot \max\{1-|{i'}_k^{(t+1)}-p|, 0 \} \cdot  \max \{1-|{j'}_k^{(t+1)}-q|, 0\} 
\end{equation}
Here, $({i'}_k^{(t+1)}, {j'}_k^{(t+1)})$ is the pixel coordinate corresponding to the predicted new 3D location ${\vm'}_{k,(i,j)}^{(t+1)}$ for pixel $(i,j)$ from the current frame. The first term $\pi_{kij}^{(t)}$ in the definition incorporates the uncertainty of which object a pixel belongs to.
The second term $e^{-\beta \cdot {{D'}_k^{(t+1)}(i,j)}}$ resolves the issue of occlusion when multiple pixels are predicted to move close to the same pixel grid by down-weighting the pixels predicted to land farther from the camera. 
These last two terms mean 
that only the source pixels predicted to land within a square of of $2 \times 2$ pixels centered at any target location $(p,q)$ will contribute to the color ${I'}_{\text{Warp}}^{(t+1)}(p,q)$.
The depth map ${\mD'}_{\text{Warp}}^{(t+1)}$ can be predicted by the same weighting scheme after replacing $I^{(t)}(i,j)$ with each predicted depth ${D'}_k^{(t+1)}(i,j)$ assuming the pixel belongs to object $k$.

Note that the formula above guarantees that the weights from all contributing original pixels sum to 1. However, because we assume that only a portion  $ \pi_{kij}^{(t)}$ of a pixel might land nearby a target pixel, the sum of all the portions of the pixels that would land in the $2\times2$ square near a target pixel may be less than 1, in which case, the above prediction is combined with the prediction based on imagination to form the final prediction for such target pixels. Therefore, the weight on warping-based prediction is calculated as $\mW_\text{warp}(p,q) = \min\{1, \pi_{kij}^{(t)}  \cdot \max\{1-|{i'}_k^{(t+1)}-p|, 0 \} \cdot  \max \{1-|{j'}_k^{(t+1)}-q|, 0\} \}$.

\section{Jointly learning rules of rigid body motion and apparent movement of objects induced by egomotion}\label{deep_warp}
As mentioned above, the expected apparent change of 3D location of a static object induced by the motion of the camera itself, and the expected movement of all pixels on an object given the object's 3D location and velocity and pixels' locations (rigid-body motion) are assumed to be known in our main model. However, it is unnatural to assume these are hard-wired in the brain. To test whether they can be jointly learned together with all other networks by optimizing for the same objective, we approximated them with neural networks. 

We used a series of 3 fully-connected layers, each with 16 units, with a residual connection from input to output, to learn the function that maps from an object's location $\hat{\vx}_{k}^{(t-1)}$, the camera's horizontal movement velocity $ \vv_\text{obs}^{(t-1)}$ and rotational speed $\omega_\text{obs}$ to the 3D location where the object would appear after the camera's motion: $\hat{x}_k^{t-1 \rightarrow t}$. We then used two convolutional networks,
each with channel sizes [64,64,32,4] and a residual connection from input to output layer, to approximate the prediction of pixels' 3D motion. One network is dedicated for the background and the other for objects. The network for predicting pixels belonging to objects takes as input a pixel's inferred 3D location, the location of the object it belongs to, the camera's motion parameters, the object's inferred 3D location and its velocity and rotational speed, and then predicts the new 3D location of the pixel in the next frame. The network for predicting background pixels' motion only takes the pixels' inferred 3D locations and camera's motion parameters as input. 

\begin{figure}[h]
\begin{center}
\includegraphics[width=1.0\textwidth]{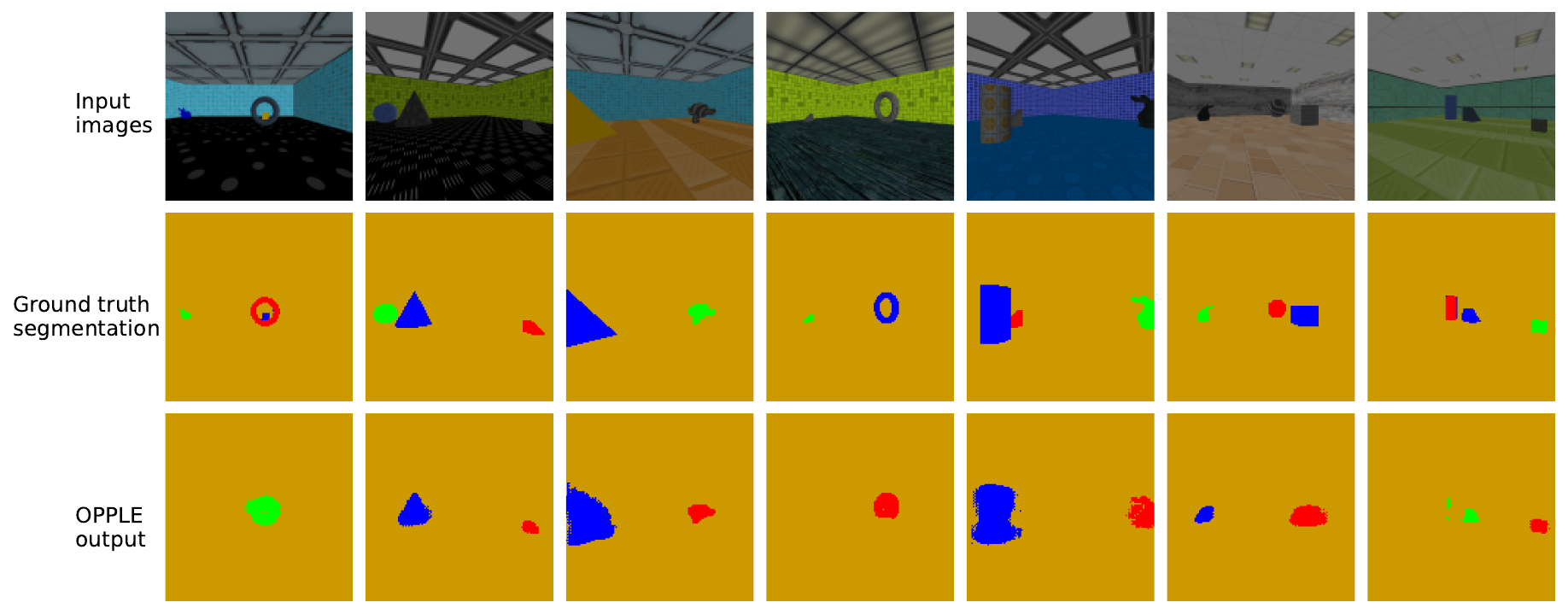}
\end{center}
\vspace{-0.2in}
\caption{Segmentation performance for relaxed model (the rule of rigid body motion and apparent object motion induced by egomotion are learned).} 
\label{fig:supp5}
\vspace{-0.1in}
\end{figure}

\begin{figure}[h]
\begin{center}
\includegraphics[width=0.8\textwidth]{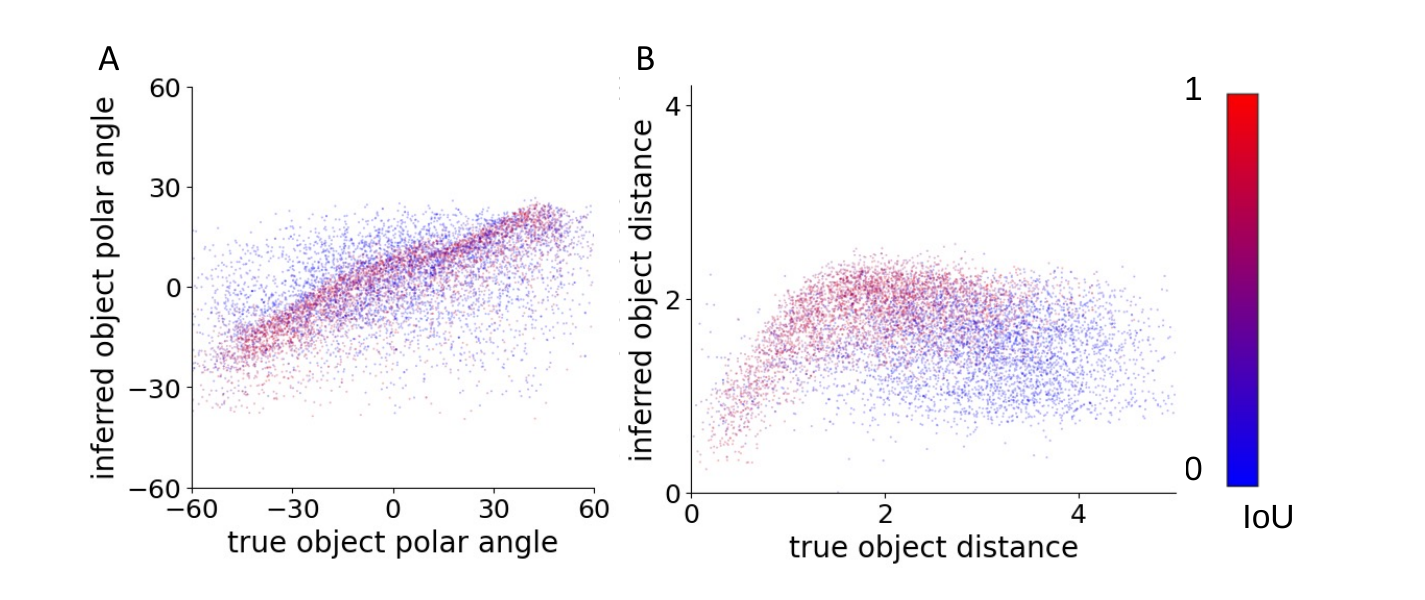}
\end{center}
\vspace{-0.2in}
\caption{Performance of object 3D localization color-coded by segmentation performance for all objects in the relaxed model.} 
\label{fig:supp6}
\vspace{-0.1in}
\end{figure}

By jointly optimizing all networks for the same objective function, we found that the relaxed model obtains a segmentation performance of ARI-fg=0.58, IoU=0.44, which is on par with the full model (ARI-fg=0.58, IoU=0.45) (Figure \ref{fig:supp5}). Further we found that the objects' 3D locations are also still inferred well when the objects are segmented well (Figure \ref{fig:supp6}). The correlation between ground truth polar angle and inferred polar angle for objects segmented with IoU>0.5 (mainly red dots) is 0.79 (while the correlation of all objects regardless of whether they can be segmented correctly by the model is 0.62). The correlation between the ground truth distance and the inferred distance for objects with IoU>0.5 is 0.48 (while the correlation without such filtering is -0.01). In regard to depth inference, we find that this network struggles to learn the background depth, which is reflected in the decrease in the overall correlation to the ground truth depth ($r=0.33$) 

As discussed in the main text, this suggests that the assumption of rigid body motion may not be strictly necessary for segmentation and 3D localization of objects together with depth perception. Note that we did make an assumption that objects move with inertia (keeping their speed) as is true in the data. Future works may test whether this assumption still allows learning all the capacities when objects actually change speed (e.g, due to collision or gravity).

\section{Performance of OPPLE on MOVi dataset} \label{appendix:MOVi-E}
The recently published MOVi dataset \citep{greff2022kubric} poses more challenges to OCRL models. It introduces more complex and realistic object motion following gravity and including interaction between objects. We therefore also test OPPLE on the most challenging version of MOVi dataset (MOVi-E) tested in a previous work SAVi++ \citep{elsayed2022savi++}. We generated scenes including 3 objects following the default data generation procedure. In order to incorporate 3D object rotation, we made modification to the object extraction network such that it outputs cosines and sines of each Euler angle characterizing the current pose $(\alpha, \beta, \gamma)$ of each object viewed by the camera. These values are used to compute a rotation matrix corresponding to the pose. Then, using the rule of apparent object movement induced by camera motion that incorporates 3D rotation, the pose of any object at $t-1$ that would appear at $t$ if it did not move is calculated. This allows the calculation of the rotation matrix for any pair of detected objects between $t-1$ and $t$. Using the same weighting introduced in Section \ref{appendix:warping_weight}, the expected rotation for each object from $t$ to $t+1$ is calculated with an assumption that the rotation from $t-1$ to $t$ carries over.

\begin{figure}[h]
\begin{center}
\includegraphics[width=1.0\textwidth]{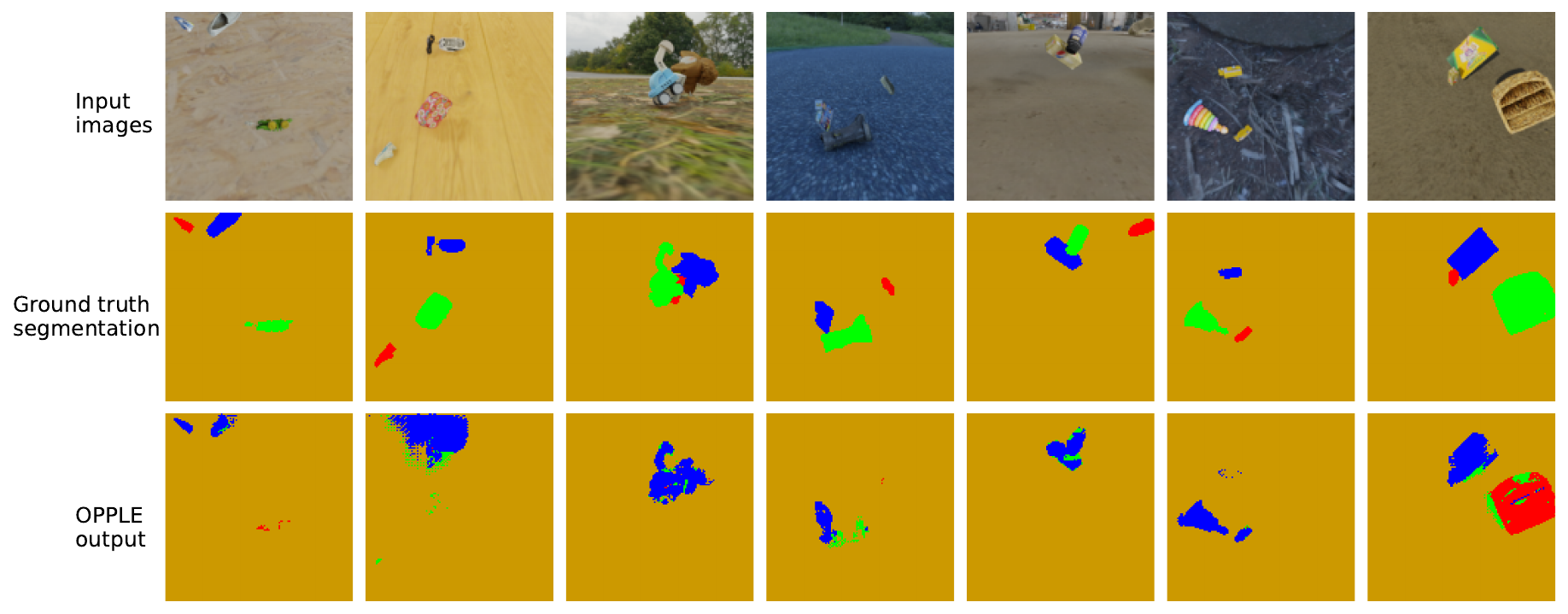}
\end{center}
\vspace{-0.2in}
\caption{OPPLE's segmentation performance on 3-object version of MOVi-E dataset.} 
\label{fig:supp7}
\vspace{-0.1in}
\end{figure}

On testing data of MOVi-E, OPPLE achieved a slightly lower segmentation performance (ARF-fg=0.38 and IoU=0.44) than on our custom dataset. Figure \ref{fig:supp7} illustrates the segmentation performance on MOVi-E dataset. It can be seen although overall the union of all segmentation masks often cover the majority of the true object masks, the model often attribute different objects into the same mask. We think the lower performance in segmentation may be due to the difficulty of learning a to assign a pose to each object relative to a canonical pose that is consistent to all objects. Additionally, because all objects follow gravitational force and fall at a similar time, their motions are often synchronized and the speed may thus appear close to each other, which encourages our network to attribute them to the same objects.

\section{Depth perception performance of O3V model on our dataset}\label{appendix:o3v}
As shown below, the O3V model generates blurrier depth maps for our dataset and often fails to capture the relative distance relationship between objects and background, even though it has the advantage of using 3 consecutive frames to infer 3D scene structure. In contrast, our model can make inference of depth based on a single image and produce more accurate depth estimation (Figure \ref{fig:supp1}). 

\begin{figure}[h]
\begin{center}
\includegraphics[width=0.8\textwidth]{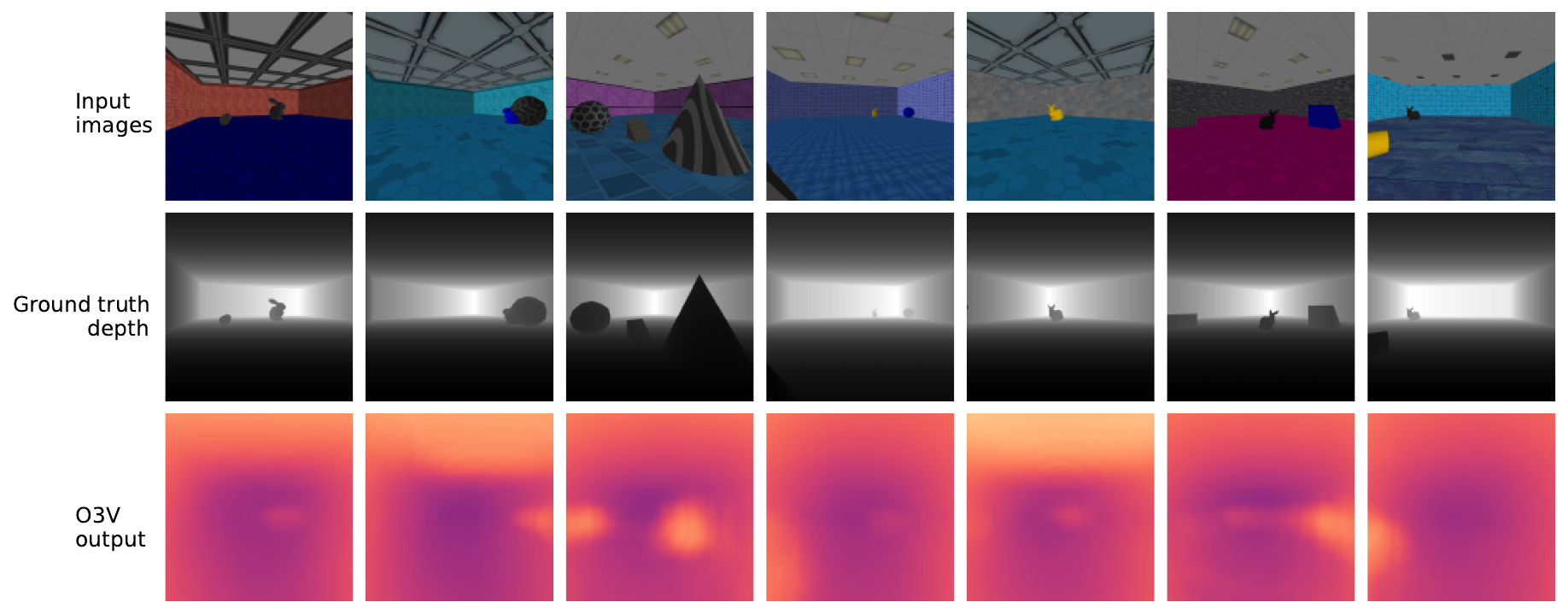}
\end{center}
\vspace{-0.2in}
\caption{Example frames (first row), the ground truth depth map (second row) and the depth map inferred by O3V (third row).} 
\label{fig:supp8}
\vspace{-0.1in}
\end{figure}

\bibliographystyle{named}
\bibliography{ijcai24}